\documentclass{article}

\usepackage[preprint]{neurips_2025}

\usepackage[utf8]{inputenc} 
\usepackage[T1]{fontenc}    
\usepackage{hyperref}       
\usepackage{url}            
\usepackage{booktabs}       
\usepackage{amsfonts}       
\usepackage{nicefrac}       
\usepackage{microtype}      
\usepackage{graphicx}
\usepackage{arydshln}
\usepackage{multirow}
\usepackage{caption}
\usepackage{comment}
\usepackage[table]{xcolor}
\usepackage{colortbl} 
\usepackage{xspace}
\usepackage{subfigure}
\usepackage[symbol]{footmisc}

\usepackage{amsmath,amsfonts,bm}

\definecolor{custom_red}{RGB}{231,111,81}
\definecolor{custom_green}{RGB}{42,157,143}
\definecolor{custom_dark}{RGB}{38,70,83}
\definecolor{custom_yellow}{RGB}{233,196,106}
\definecolor{custom_orange}{RGB}{244,162,97}

\DeclareMathAlphabet{\mathsfit}{\encodingdefault}{\sfdefault}{m}{sl}
\SetMathAlphabet{\mathsfit}{bold}{\encodingdefault}{\sfdefault}{bx}{n}

\title{\Approach: Accelerating Dense 3D Tracking}

\author{Tuan Duc Ngo$^\dagger$\\ UMass Amherst \And Ashkan Mirzaei \\ Snap Inc.  \And Guocheng Qian \\ Snap Inc. \And Hanwen Liang$^\dagger$ \\ University of Toronto \And Chuang Gan \\ UMass Amherst  \And Evangelos Kalogerakis \\ UMass Amherst \& TU Crete   \And Peter Wonka \\ Snap Inc. \& KAUST \And Chaoyang Wang \\ Snap Inc.}

\begin{document}
\def\mA{\mathcal{A}}
\def\mB{\mathcal{B}}
\def\mC{\mathcal{C}}
\def\mD{\mathcal{D}}
\def\mE{\mathcal{E}}
\def\mF{\mathcal{F}}
\def\mG{\mathcal{G}}
\def\mH{\mathcal{H}}
\def\mI{\mathcal{I}}
\def\mJ{\mathcal{J}}
\def\mK{\mathcal{K}}
\def\mL{\mathcal{L}}
\def\mM{\mathcal{M}}
\def\mN{\mathcal{N}}
\def\mO{\mathcal{O}}
\def\mP{\mathcal{P}}
\def\mQ{\mathcal{Q}}
\def\mR{\mathcal{R}}
\def\mS{\mathcal{S}}
\def\mT{\mathcal{T}}
\def\mU{\mathcal{U}}
\def\mV{\mathcal{V}}
\def\mW{\mathcal{W}}
\def\mX{\mathcal{X}}
\def\mY{\mathcal{Y}}
\def\mZ{\mathcal{Z}} 

\def\bbN{\mathbb{N}} 
\def\bbR{\mathbb{R}} 
\def\bbP{\mathbb{P}} 
\def\bbQ{\mathbb{Q}} 
\def\bbE{\mathbb{E}}

\def\1n{\mathbf{1}_n}
\def\0{\mathbf{0}}
\def\1{\mathbf{1}}

\def\A{{\bf A}}
\def\B{{\bf B}}
\def\C{{\bf C}}
\def\D{{\bf D}}
\def\E{{\bf E}}
\def\F{{\bf F}}
\def\G{{\bf G}}
\def\H{{\bf H}}
\def\I{{\bf I}}
\def\J{{\bf J}}
\def\K{{\bf K}}
\def\L{{\bf L}}
\def\M{{\bf M}}
\def\N{{\bf N}}
\def\O{{\bf O}}
\def\P{{\bf P}}
\def\Q{{\bf Q}}
\def\R{{\bf R}}
\def\S{{\bf S}}
\def\T{{\bf T}}
\def\U{{\bf U}}
\def\V{{\bf V}}
\def\W{{\bf W}}
\def\X{{\bf X}}
\def\Y{{\bf Y}}
\def\Z{{\bf Z}}

\def\a{{\bf a}}
\def\b{{\bf b}}
\def\c{{\bf c}}
\def\d{{\bf d}}
\def\e{{\bf e}}
\def\f{{\bf f}}
\def\g{{\bf g}}
\def\h{{\bf h}}
\def\i{{\bf i}}
\def\j{{\bf j}}
\def\k{{\bf k}}
\def\l{{\bf l}}
\def\m{{\bf m}}
\def\n{{\bf n}}
\def\o{{\bf o}}
\def\p{{\bf p}}
\def\q{{\bf q}}
\def\r{{\bf r}}
\def\s{{\bf s}}
\def\t{{\bf t}}
\def\u{{\bf u}}
\def\v{{\bf v}}
\def\w{{\bf w}}
\def\x{{\bf x}}
\def\y{{\bf y}}
\def\z{{\bf z}}

\def\balpha{\mbox{\boldmath{$\alpha$}}}
\def\bbeta{\mbox{\boldmath{$\beta$}}}
\def\bdelta{\mbox{\boldmath{$\delta$}}}
\def\bgamma{\mbox{\boldmath{$\gamma$}}}
\def\blambda{\mbox{\boldmath{$\lambda$}}}
\def\bsigma{\mbox{\boldmath{$\sigma$}}}
\def\btheta{\mbox{\boldmath{$\theta$}}}
\def\bomega{\mbox{\boldmath{$\omega$}}}
\def\bxi{\mbox{\boldmath{$\xi$}}}
\def\bnu{\mbox{\boldmath{$\nu$}}}                                  
\def\bphi{\mbox{\boldmath{$\phi$}}}
\def\bmu{\mbox{\boldmath{$\mu$}}}

\def\bDelta{\mbox{\boldmath{$\Delta$}}}
\def\bOmega{\mbox{\boldmath{$\Omega$}}}
\def\bPhi{\mbox{\boldmath{$\Phi$}}}
\def\bLambda{\mbox{\boldmath{$\Lambda$}}}
\def\bSigma{\mbox{\boldmath{$\Sigma$}}}
\def\bGamma{\mbox{\boldmath{$\Gamma$}}}
                                  
\newcommand{\myprob}[1]{\mathop{\mathbb{P}}_{#1}}

\newcommand{\myexp}[1]{\mathop{\mathbb{E}}_{#1}}

\newcommand{\mydelta}[1]{1_{#1}}

\newcommand{\myminimum}[1]{\mathop{\textrm{minimum}}_{#1}}
\newcommand{\mymaximum}[1]{\mathop{\textrm{maximum}}_{#1}}    
\newcommand{\mymin}[1]{\mathop{\textrm{minimize}}_{#1}}
\newcommand{\mymax}[1]{\mathop{\textrm{maximize}}_{#1}}
\newcommand{\mymins}[1]{\mathop{\textrm{min.}}_{#1}}
\newcommand{\mymaxs}[1]{\mathop{\textrm{max.}}_{#1}}  
\newcommand{\myargmin}[1]{\mathop{\textrm{argmin}}_{#1}} 
\newcommand{\myargmax}[1]{\mathop{\textrm{argmax}}_{#1}} 
\newcommand{\myst}{\textrm{s.t. }}

\newcommand{\denselist}{\itemsep -1pt}
\newcommand{\sparselist}{\itemsep 1pt}

\definecolor{pink}{rgb}{0.9,0.5,0.5}
\definecolor{purple}{rgb}{0.5, 0.4, 0.8}   
\definecolor{gray}{rgb}{0.3, 0.3, 0.3}
\definecolor{mygreen}{rgb}{0.2, 0.6, 0.2}

\newcommand{\cyan}[1]{\textcolor{cyan}{#1}}
\newcommand{\red}[1]{\textcolor{red}{#1}}  
\newcommand{\blue}[1]{\textcolor{blue}{#1}}
\newcommand{\magenta}[1]{\textcolor{magenta}{#1}}
\newcommand{\pink}[1]{\textcolor{pink}{#1}}
\newcommand{\green}[1]{\textcolor{green}{#1}} 
\newcommand{\gray}[1]{\textcolor{gray}{#1}}    
\newcommand{\mygreen}[1]{\textcolor{mygreen}{#1}}    
\newcommand{\purple}[1]{\textcolor{purple}{#1}}       

\definecolor{greena}{rgb}{0.4, 0.5, 0.1}
\newcommand{\greena}[1]{\textcolor{greena}{#1}}

\definecolor{bluea}{rgb}{0, 0.4, 0.6}
\newcommand{\bluea}[1]{\textcolor{bluea}{#1}}
\definecolor{reda}{rgb}{0.6, 0.2, 0.1}
\newcommand{\reda}[1]{\textcolor{reda}{#1}}

\def\changemargin#1#2{\list{}{\rightmargin#2\leftmargin#1}\item[]}
\let\endchangemargin=\endlist
                                               
\newcommand{\cm}[1]{}

\newcommand{\mhoai}[1]{{\color{magenta}\textbf{[MH: #1]}}}

\newcommand{\mtodo}[1]{{\color{red}$\blacksquare$\textbf{[TODO: #1]}}}
\newcommand{\myheading}[1]{\vspace{1ex}\noindent \textbf{#1}}
\newcommand{\htimesw}[2]{\mbox{$#1$$\times$$#2$}}


\newif\ifshowsolution
\showsolutiontrue

\ifshowsolution  
\newcommand{\Solution}[2]{\paragraph{\bf $\bigstar $ SOLUTION:} {\sf #2} }
\newcommand{\Mistake}[2]{\paragraph{\bf $\blacksquare$ COMMON MISTAKE #1:} {\sf #2} \bigskip}
\else
\newcommand{\Solution}[2]{\vspace{#1}}
\fi

\newcommand{\truefalse}{
\begin{enumerate}
	\item True
	\item False
\end{enumerate}
}

\newcommand{\yesno}{
\begin{enumerate}
	\item Yes
	\item No
\end{enumerate}
}

\newcommand{\Sref}[1]{Sec.~\ref{#1}}
\newcommand{\Eref}[1]{Eq.~(\ref{#1})}
\newcommand{\Fref}[1]{Fig.~\ref{#1}}
\newcommand{\Tref}[1]{Table~\ref{#1}}

\newcommand\blfootnote[1]{%
  \begingroup
  \renewcommand\thefootnote{}\footnote{#1}%
  \addtocounter{footnote}{-1}%
  \endgroup
}

\makeatletter
\DeclareRobustCommand\onedot{\futurelet\@let@token\@onedot}
\def\@onedot{\ifx\@let@token.\else.\null\fi\xspace}

\newcommand{\eg}{\emph{e.g}\onedot}
\newcommand{\Eg}{\emph{E.g}\onedot}
\newcommand{\ie}{\emph{i.e}\onedot}
\newcommand{\Ie}{\emph{I.e}\onedot}
\newcommand{\cf}{\emph{cf}\onedot}
\newcommand{\Cf}{\emph{Cf}\onedot}
\newcommand{\etc}{\emph{etc}\onedot}
\newcommand{\wrt}{w.r.t\onedot}
\newcommand{\dof}{d.o.f\onedot}
\newcommand{\iid}{i.i.d\onedot}
\newcommand{\wolog}{w.l.o.g\onedot}
\newcommand{\etal}{\emph{et al}\onedot}

\def\paperID{xxxxx} 
\def\confName{NIPS}
\def\confYear{2025}

\def\Approach{DELTAv2\xspace}

\newcommand{\gqc}[1]{{\color{orange}[\textbf{Gordon:} #1]}}
\newcommand{\gq}[1]{{\color{orange}#1}}
\newcommand{\cy}[1]{{\color{blue}[\textbf{CY:} #1]}}
\newcommand{\tu}[1]{{\color{green}[\textbf{Tuan:} #1]}}
\newcommand{\am}[1]{{\color{red}[\textbf{Ashkan:} #1]}}

\definecolor{graybg}{gray}{0.75} 

\definecolor{custom_red}{RGB}{231,111,81}
\definecolor{custom_green}{RGB}{42,157,143}
\definecolor{custom_dark}{RGB}{38,70,83}
\definecolor{custom_yellow}{RGB}{233,196,106}
\definecolor{custom_orange}{RGB}{244,162,97}


\newcommand{\figleft}{{\em (Left)}}
\newcommand{\figcenter}{{\em (Center)}}
\newcommand{\figright}{{\em (Right)}}
\newcommand{\figtop}{{\em (Top)}}
\newcommand{\figbottom}{{\em (Bottom)}}
\newcommand{\captiona}{{\em (a)}}
\newcommand{\captionb}{{\em (b)}}
\newcommand{\captionc}{{\em (c)}}
\newcommand{\captiond}{{\em (d)}}

\newcommand{\newterm}[1]{{\bf #1}}

\def\figref#1{figure~\ref{#1}}
\def\Figref#1{Figure~\ref{#1}}
\def\twofigref#1#2{figures \ref{#1} and \ref{#2}}
\def\quadfigref#1#2#3#4{figures \ref{#1}, \ref{#2}, \ref{#3} and \ref{#4}}
\def\secref#1{section~\ref{#1}}
\def\Secref#1{Section~\ref{#1}}
\def\twosecrefs#1#2{sections \ref{#1} and \ref{#2}}
\def\secrefs#1#2#3{sections \ref{#1}, \ref{#2} and \ref{#3}}
\def\eqref#1{equation~\ref{#1}}
\def\Eqref#1{Equation~\ref{#1}}
\def\plaineqref#1{\ref{#1}}
\def\chapref#1{chapter~\ref{#1}}
\def\Chapref#1{Chapter~\ref{#1}}
\def\rangechapref#1#2{chapters\ref{#1}--\ref{#2}}
\def\algref#1{algorithm~\ref{#1}}
\def\Algref#1{Algorithm~\ref{#1}}
\def\twoalgref#1#2{algorithms \ref{#1} and \ref{#2}}
\def\Twoalgref#1#2{Algorithms \ref{#1} and \ref{#2}}
\def\partref#1{part~\ref{#1}}
\def\Partref#1{Part~\ref{#1}}
\def\twopartref#1#2{parts \ref{#1} and \ref{#2}}

\def\ceil#1{\lceil #1 \rceil}
\def\floor#1{\lfloor #1 \rfloor}
\def\1{\bm{1}}
\newcommand{\train}{\mathcal{D}}
\newcommand{\valid}{\mathcal{D_{\mathrm{valid}}}}
\newcommand{\test}{\mathcal{D_{\mathrm{test}}}}

\def\eps{{\epsilon}}

\def\reta{{\textnormal{$\eta$}}}
\def\ra{{\textnormal{a}}}
\def\rb{{\textnormal{b}}}
\def\rc{{\textnormal{c}}}
\def\rd{{\textnormal{d}}}
\def\re{{\textnormal{e}}}
\def\rf{{\textnormal{f}}}
\def\rg{{\textnormal{g}}}
\def\rh{{\textnormal{h}}}
\def\ri{{\textnormal{i}}}
\def\rj{{\textnormal{j}}}
\def\rk{{\textnormal{k}}}
\def\rl{{\textnormal{l}}}
\def\rn{{\textnormal{n}}}
\def\ro{{\textnormal{o}}}
\def\rp{{\textnormal{p}}}
\def\rq{{\textnormal{q}}}
\def\rr{{\textnormal{r}}}
\def\rs{{\textnormal{s}}}
\def\rt{{\textnormal{t}}}
\def\ru{{\textnormal{u}}}
\def\rv{{\textnormal{v}}}
\def\rw{{\textnormal{w}}}
\def\rx{{\textnormal{x}}}
\def\ry{{\textnormal{y}}}
\def\rz{{\textnormal{z}}}

\def\rvepsilon{{\mathbf{\epsilon}}}
\def\rvtheta{{\mathbf{\theta}}}
\def\rva{{\mathbf{a}}}
\def\rvb{{\mathbf{b}}}
\def\rvc{{\mathbf{c}}}
\def\rvd{{\mathbf{d}}}
\def\rve{{\mathbf{e}}}
\def\rvf{{\mathbf{f}}}
\def\rvg{{\mathbf{g}}}
\def\rvh{{\mathbf{h}}}
\def\rvu{{\mathbf{i}}}
\def\rvj{{\mathbf{j}}}
\def\rvk{{\mathbf{k}}}
\def\rvl{{\mathbf{l}}}
\def\rvm{{\mathbf{m}}}
\def\rvn{{\mathbf{n}}}
\def\rvo{{\mathbf{o}}}
\def\rvp{{\mathbf{p}}}
\def\rvq{{\mathbf{q}}}
\def\rvr{{\mathbf{r}}}
\def\rvs{{\mathbf{s}}}
\def\rvt{{\mathbf{t}}}
\def\rvu{{\mathbf{u}}}
\def\rvv{{\mathbf{v}}}
\def\rvw{{\mathbf{w}}}
\def\rvx{{\mathbf{x}}}
\def\rvy{{\mathbf{y}}}
\def\rvz{{\mathbf{z}}}

\def\erva{{\textnormal{a}}}
\def\ervb{{\textnormal{b}}}
\def\ervc{{\textnormal{c}}}
\def\ervd{{\textnormal{d}}}
\def\erve{{\textnormal{e}}}
\def\ervf{{\textnormal{f}}}
\def\ervg{{\textnormal{g}}}
\def\ervh{{\textnormal{h}}}
\def\ervi{{\textnormal{i}}}
\def\ervj{{\textnormal{j}}}
\def\ervk{{\textnormal{k}}}
\def\ervl{{\textnormal{l}}}
\def\ervm{{\textnormal{m}}}
\def\ervn{{\textnormal{n}}}
\def\ervo{{\textnormal{o}}}
\def\ervp{{\textnormal{p}}}
\def\ervq{{\textnormal{q}}}
\def\ervr{{\textnormal{r}}}
\def\ervs{{\textnormal{s}}}
\def\ervt{{\textnormal{t}}}
\def\ervu{{\textnormal{u}}}
\def\ervv{{\textnormal{v}}}
\def\ervw{{\textnormal{w}}}
\def\ervx{{\textnormal{x}}}
\def\ervy{{\textnormal{y}}}
\def\ervz{{\textnormal{z}}}

\def\rmA{{\mathbf{A}}}
\def\rmB{{\mathbf{B}}}
\def\rmC{{\mathbf{C}}}
\def\rmD{{\mathbf{D}}}
\def\rmE{{\mathbf{E}}}
\def\rmF{{\mathbf{F}}}
\def\rmG{{\mathbf{G}}}
\def\rmH{{\mathbf{H}}}
\def\rmI{{\mathbf{I}}}
\def\rmJ{{\mathbf{J}}}
\def\rmK{{\mathbf{K}}}
\def\rmL{{\mathbf{L}}}
\def\rmM{{\mathbf{M}}}
\def\rmN{{\mathbf{N}}}
\def\rmO{{\mathbf{O}}}
\def\rmP{{\mathbf{P}}}
\def\rmQ{{\mathbf{Q}}}
\def\rmR{{\mathbf{R}}}
\def\rmS{{\mathbf{S}}}
\def\rmT{{\mathbf{T}}}
\def\rmU{{\mathbf{U}}}
\def\rmV{{\mathbf{V}}}
\def\rmW{{\mathbf{W}}}
\def\rmX{{\mathbf{X}}}
\def\rmY{{\mathbf{Y}}}
\def\rmZ{{\mathbf{Z}}}

\def\ermA{{\textnormal{A}}}
\def\ermB{{\textnormal{B}}}
\def\ermC{{\textnormal{C}}}
\def\ermD{{\textnormal{D}}}
\def\ermE{{\textnormal{E}}}
\def\ermF{{\textnormal{F}}}
\def\ermG{{\textnormal{G}}}
\def\ermH{{\textnormal{H}}}
\def\ermI{{\textnormal{I}}}
\def\ermJ{{\textnormal{J}}}
\def\ermK{{\textnormal{K}}}
\def\ermL{{\textnormal{L}}}
\def\ermM{{\textnormal{M}}}
\def\ermN{{\textnormal{N}}}
\def\ermO{{\textnormal{O}}}
\def\ermP{{\textnormal{P}}}
\def\ermQ{{\textnormal{Q}}}
\def\ermR{{\textnormal{R}}}
\def\ermS{{\textnormal{S}}}
\def\ermT{{\textnormal{T}}}
\def\ermU{{\textnormal{U}}}
\def\ermV{{\textnormal{V}}}
\def\ermW{{\textnormal{W}}}
\def\ermX{{\textnormal{X}}}
\def\ermY{{\textnormal{Y}}}
\def\ermZ{{\textnormal{Z}}}

\def\vzero{{\bm{0}}}
\def\vone{{\bm{1}}}
\def\vmu{{\bm{\mu}}}
\def\vtheta{{\bm{\theta}}}
\def\va{{\bm{a}}}
\def\vb{{\bm{b}}}
\def\vc{{\bm{c}}}
\def\vd{{\bm{d}}}
\def\ve{{\bm{e}}}
\def\vf{{\bm{f}}}
\def\vg{{\bm{g}}}
\def\vh{{\bm{h}}}
\def\vi{{\bm{i}}}
\def\vj{{\bm{j}}}
\def\vk{{\bm{k}}}
\def\vl{{\bm{l}}}
\def\vm{{\bm{m}}}
\def\vn{{\bm{n}}}
\def\vo{{\bm{o}}}
\def\vp{{\bm{p}}}
\def\vq{{\bm{q}}}
\def\vr{{\bm{r}}}
\def\vs{{\bm{s}}}
\def\vt{{\bm{t}}}
\def\vu{{\bm{u}}}
\def\vv{{\bm{v}}}
\def\vw{{\bm{w}}}
\def\vx{{\bm{x}}}
\def\vy{{\bm{y}}}
\def\vz{{\bm{z}}}

\def\evalpha{{\alpha}}
\def\evbeta{{\beta}}
\def\evepsilon{{\epsilon}}
\def\evlambda{{\lambda}}
\def\evomega{{\omega}}
\def\evmu{{\mu}}
\def\evpsi{{\psi}}
\def\evsigma{{\sigma}}
\def\evtheta{{\theta}}
\def\eva{{a}}
\def\evb{{b}}
\def\evc{{c}}
\def\evd{{d}}
\def\eve{{e}}
\def\evf{{f}}
\def\evg{{g}}
\def\evh{{h}}
\def\evi{{i}}
\def\evj{{j}}
\def\evk{{k}}
\def\evl{{l}}
\def\evm{{m}}
\def\evn{{n}}
\def\evo{{o}}
\def\evp{{p}}
\def\evq{{q}}
\def\evr{{r}}
\def\evs{{s}}
\def\evt{{t}}
\def\evu{{u}}
\def\evv{{v}}
\def\evw{{w}}
\def\evx{{x}}
\def\evy{{y}}
\def\evz{{z}}

\def\mA{{\bm{A}}}
\def\mB{{\bm{B}}}
\def\mC{{\bm{C}}}
\def\mD{{\bm{D}}}
\def\mE{{\bm{E}}}
\def\mF{{\bm{F}}}
\def\mG{{\bm{G}}}
\def\mH{{\bm{H}}}
\def\mI{{\bm{I}}}
\def\mJ{{\bm{J}}}
\def\mK{{\bm{K}}}
\def\mL{{\bm{L}}}
\def\mM{{\bm{M}}}
\def\mN{{\bm{N}}}
\def\mO{{\bm{O}}}
\def\mP{{\bm{P}}}
\def\mQ{{\bm{Q}}}
\def\mR{{\bm{R}}}
\def\mS{{\bm{S}}}
\def\mT{{\bm{T}}}
\def\mU{{\bm{U}}}
\def\mV{{\bm{V}}}
\def\mW{{\bm{W}}}
\def\mX{{\bm{X}}}
\def\mY{{\bm{Y}}}
\def\mZ{{\bm{Z}}}
\def\mBeta{{\bm{\beta}}}
\def\mPhi{{\bm{\Phi}}}
\def\mLambda{{\bm{\Lambda}}}
\def\mSigma{{\bm{\Sigma}}}

\newcommand{\tens}[1]{\bm{\mathsfit{#1}}}
\def\tA{{\tens{A}}}
\def\tB{{\tens{B}}}
\def\tC{{\tens{C}}}
\def\tD{{\tens{D}}}
\def\tE{{\tens{E}}}
\def\tF{{\tens{F}}}
\def\tG{{\tens{G}}}
\def\tH{{\tens{H}}}
\def\tI{{\tens{I}}}
\def\tJ{{\tens{J}}}
\def\tK{{\tens{K}}}
\def\tL{{\tens{L}}}
\def\tM{{\tens{M}}}
\def\tN{{\tens{N}}}
\def\tO{{\tens{O}}}
\def\tP{{\tens{P}}}
\def\tQ{{\tens{Q}}}
\def\tR{{\tens{R}}}
\def\tS{{\tens{S}}}
\def\tT{{\tens{T}}}
\def\tU{{\tens{U}}}
\def\tV{{\tens{V}}}
\def\tW{{\tens{W}}}
\def\tX{{\tens{X}}}
\def\tY{{\tens{Y}}}
\def\tZ{{\tens{Z}}}

\def\gA{{\mathcal{A}}}
\def\gB{{\mathcal{B}}}
\def\gC{{\mathcal{C}}}
\def\gD{{\mathcal{D}}}
\def\gE{{\mathcal{E}}}
\def\gF{{\mathcal{F}}}
\def\gG{{\mathcal{G}}}
\def\gH{{\mathcal{H}}}
\def\gI{{\mathcal{I}}}
\def\gJ{{\mathcal{J}}}
\def\gK{{\mathcal{K}}}
\def\gL{{\mathcal{L}}}
\def\gM{{\mathcal{M}}}
\def\gN{{\mathcal{N}}}
\def\gO{{\mathcal{O}}}
\def\gP{{\mathcal{P}}}
\def\gQ{{\mathcal{Q}}}
\def\gR{{\mathcal{R}}}
\def\gS{{\mathcal{S}}}
\def\gT{{\mathcal{T}}}
\def\gU{{\mathcal{U}}}
\def\gV{{\mathcal{V}}}
\def\gW{{\mathcal{W}}}
\def\gX{{\mathcal{X}}}
\def\gY{{\mathcal{Y}}}
\def\gZ{{\mathcal{Z}}}

\def\sA{{\mathbb{A}}}
\def\sB{{\mathbb{B}}}
\def\sC{{\mathbb{C}}}
\def\sD{{\mathbb{D}}}
\def\sF{{\mathbb{F}}}
\def\sG{{\mathbb{G}}}
\def\sH{{\mathbb{H}}}
\def\sI{{\mathbb{I}}}
\def\sJ{{\mathbb{J}}}
\def\sK{{\mathbb{K}}}
\def\sL{{\mathbb{L}}}
\def\sM{{\mathbb{M}}}
\def\sN{{\mathbb{N}}}
\def\sO{{\mathbb{O}}}
\def\sP{{\mathbb{P}}}
\def\sQ{{\mathbb{Q}}}
\def\sR{{\mathbb{R}}}
\def\sS{{\mathbb{S}}}
\def\sT{{\mathbb{T}}}
\def\sU{{\mathbb{U}}}
\def\sV{{\mathbb{V}}}
\def\sW{{\mathbb{W}}}
\def\sX{{\mathbb{X}}}
\def\sY{{\mathbb{Y}}}
\def\sZ{{\mathbb{Z}}}

\def\emLambda{{\Lambda}}
\def\emA{{A}}
\def\emB{{B}}
\def\emC{{C}}
\def\emD{{D}}
\def\emE{{E}}
\def\emF{{F}}
\def\emG{{G}}
\def\emH{{H}}
\def\emI{{I}}
\def\emJ{{J}}
\def\emK{{K}}
\def\emL{{L}}
\def\emM{{M}}
\def\emN{{N}}
\def\emO{{O}}
\def\emP{{P}}
\def\emQ{{Q}}
\def\emR{{R}}
\def\emS{{S}}
\def\emT{{T}}
\def\emU{{U}}
\def\emV{{V}}
\def\emW{{W}}
\def\emX{{X}}
\def\emY{{Y}}
\def\emZ{{Z}}
\def\emSigma{{\Sigma}}

\newcommand{\etens}[1]{\mathsfit{#1}}
\def\etLambda{{\etens{\Lambda}}}
\def\etA{{\etens{A}}}
\def\etB{{\etens{B}}}
\def\etC{{\etens{C}}}
\def\etD{{\etens{D}}}
\def\etE{{\etens{E}}}
\def\etF{{\etens{F}}}
\def\etG{{\etens{G}}}
\def\etH{{\etens{H}}}
\def\etI{{\etens{I}}}
\def\etJ{{\etens{J}}}
\def\etK{{\etens{K}}}
\def\etL{{\etens{L}}}
\def\etM{{\etens{M}}}
\def\etN{{\etens{N}}}
\def\etO{{\etens{O}}}
\def\etP{{\etens{P}}}
\def\etQ{{\etens{Q}}}
\def\etR{{\etens{R}}}
\def\etS{{\etens{S}}}
\def\etT{{\etens{T}}}
\def\etU{{\etens{U}}}
\def\etV{{\etens{V}}}
\def\etW{{\etens{W}}}
\def\etX{{\etens{X}}}
\def\etY{{\etens{Y}}}
\def\etZ{{\etens{Z}}}

\newcommand{\pdata}{p_{\rm{data}}}
\newcommand{\ptrain}{\hat{p}_{\rm{data}}}
\newcommand{\Ptrain}{\hat{P}_{\rm{data}}}
\newcommand{\pmodel}{p_{\rm{model}}}
\newcommand{\Pmodel}{P_{\rm{model}}}
\newcommand{\ptildemodel}{\tilde{p}_{\rm{model}}}
\newcommand{\pencode}{p_{\rm{encoder}}}
\newcommand{\pdecode}{p_{\rm{decoder}}}
\newcommand{\precons}{p_{\rm{reconstruct}}}

\newcommand{\laplace}{\mathrm{Laplace}} 

\newcommand{\Ls}{\mathcal{L}}
\newcommand{\emp}{\tilde{p}}
\newcommand{\lr}{\alpha}
\newcommand{\reg}{\lambda}
\newcommand{\rect}{\mathrm{rectifier}}
\newcommand{\softmax}{\mathrm{softmax}}
\newcommand{\sigmoid}{\sigma}
\newcommand{\softplus}{\zeta}
\newcommand{\KL}{D_{\mathrm{KL}}}
\newcommand{\Var}{\mathrm{Var}}
\newcommand{\standarderror}{\mathrm{SE}}
\newcommand{\Cov}{\mathrm{Cov}}
\newcommand{\normlzero}{L^0}
\newcommand{\normlone}{L^1}
\newcommand{\normltwo}{L^2}
\newcommand{\normlp}{L^p}
\newcommand{\normmax}{L^\infty}

\newcommand{\parents}{Pa} 

\let\ab\allowbreak

\maketitle

\vspace{-25pt}
\begin{center}
{\url{https://snap-research.github.io/DELTAv2/}}
\end{center}
\vspace{15pt}

\begin{figure}[th]
\begin{center}
\vspace{-8mm}
\includegraphics[width=0.98\linewidth]{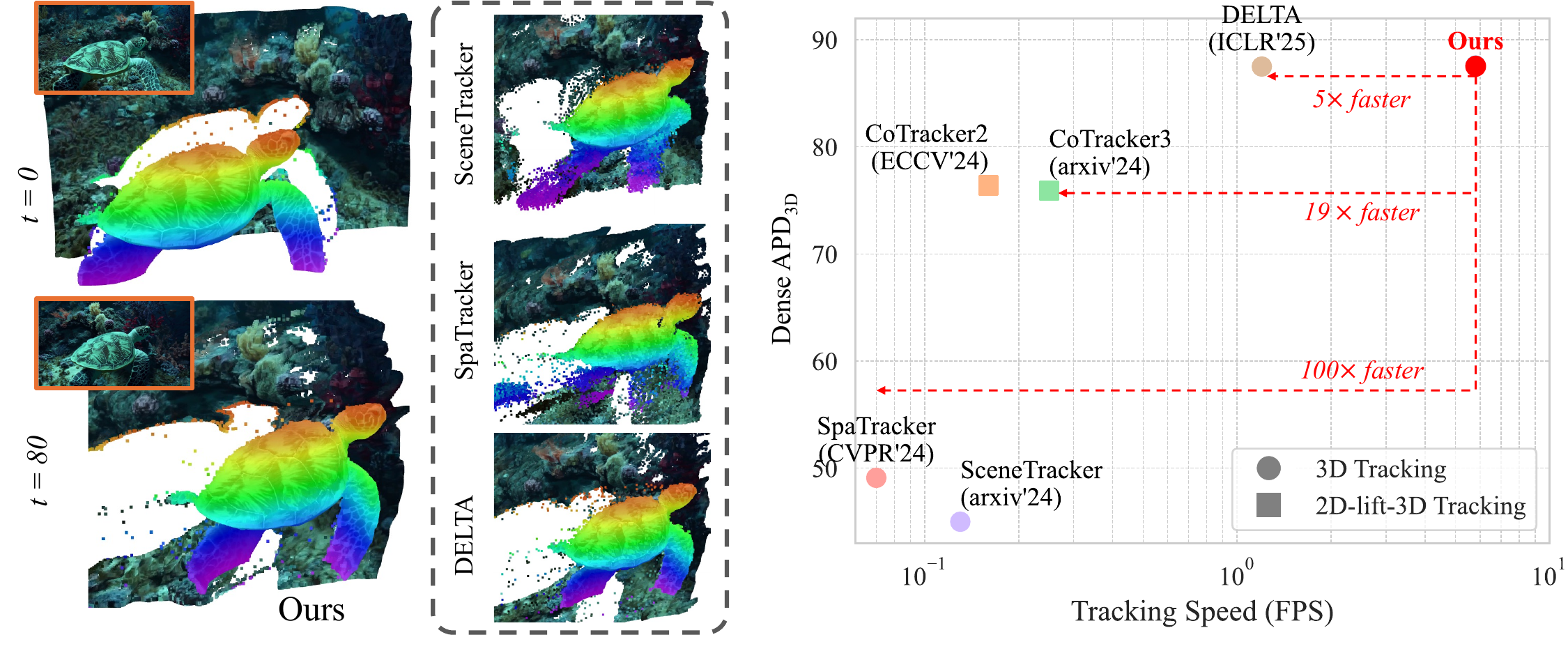}
\vspace{-2mm}
\captionof{figure}{\textbf{\Approach} achieves state-of-the-art for dense 3D tracking — matching the accuracy of DELTA while being $5\times$ faster, and outperforming all prior methods in speed–accuracy tradeoff. (Left) Long-range 3D trajectories on real-world videos. (Right) Performance vs. FPS comparison.
}
\label{fig:teaser}
\end{center}
\vspace{-4mm}
\end{figure}

\vspace{-1mm}
\begin{abstract}
We propose a novel algorithm for accelerating dense long-term 3D point tracking in videos. Through analysis of existing state-of-the-art methods, we identify two major computational bottlenecks. First, transformer-based iterative tracking becomes expensive when handling a large number of trajectories. To address this, we introduce a coarse-to-fine strategy that begins tracking with a small subset of points and progressively expands the set of tracked trajectories. The newly added trajectories are initialized using a learnable interpolation module, which is trained end-to-end alongside the tracking network. Second, we propose an optimization that significantly reduces the cost of correlation feature computation, another key bottleneck in prior methods. Together, these improvements lead to a 5–100$\times$ speedup over existing approaches while maintaining state-of-the-art tracking accuracy.

\end{abstract}
\vspace{-2mm}

\section{Introduction}
\label{sec:intro}
\begingroup
\footnotetext{$^{\dagger}$Work done during internship at Snap Inc.}
\endgroup

We focus on \emph{dense 3D tracking}: given an video, we predict the 3D trajectory of every pixel in the first frame across all subsequent frames in the local camera coordinate system. This task is more challenging than related problems such as optical flow~\cite{memin1998dense,horn1980determining,brox2004high,flownet,ranjan2017optical,xu2017accurate,pwcnet,raft,flowformer,gmflow} and scene flow~\cite{hadfield2011kinecting,hornacek2014sphereflow,quiroga2014dense,raft3d,flownet3d,flownet3d++,niemeyer2019occupancy}, which estimate motion only between adjacent frames. In contrast, dense 3D tracking requires maintaining accurate long-term correspondences under larger motions and appearance changes.

To achieve long-term point tracking, recent methods leverage powerful transformer-based networks to estimate trajectories from videos~\cite{pips,tapir,taptr,taptrv2,cotracker,cotracker3}. Some works~\cite{spatialtracker,scenetracker} further extend tracking from 2D to 3D by incorporating accurate monocular depth estimation~\cite{zoedepth,unidepth,depthcrafter}. However, these approaches remain limited to sparse trajectories, short temporal windows, or both.

DELTA~\cite{ngo2024delta} addressed these limitations by introducing the first framework for dense, long-range 3D tracking. While DELTA achieves strong performance and is relatively fast compared to prior work, it remains computationally expensive: tracking every pixel in a 100-frame video takes around 2 minutes, too slow for real-time or latency-sensitive applications. 

In this paper, we present a faster and more scalable successor to DELTA. We begin by identifying two key computational bottlenecks in the DELTA pipeline. 

First, despite the efficient transformer design, its iterative refinement must process a large number of trajectory tokens repeatedly for every tracking iteration, leading to high computational cost.  We observe that in a dense motion field, many nearby pixels exhibit similar motion patterns. As such, tracking all of them through every iteration is often redundant. Motivated by this, we analyze three possible coarse-to-fine strategies and propose an algorithm that reduces computation by subsampling a sparse set of trajectories in early iterations. The tracking density is then progressively increased in later iterations, culminating in fully dense tracking in the final iteration to ensure accuracy is preserved. To recover the motion of untracked pixels during intermediate iterations, we introduce a learnable interpolation module. This module dynamically predicts blending weights to infer the motion of untracked pixels from their nearby tracked neighbors, leveraging both spatial and feature similarity. 

\begin{figure*}[t]
\begin{minipage}{0.43\linewidth}
    \centering
    \includegraphics[width=\linewidth]{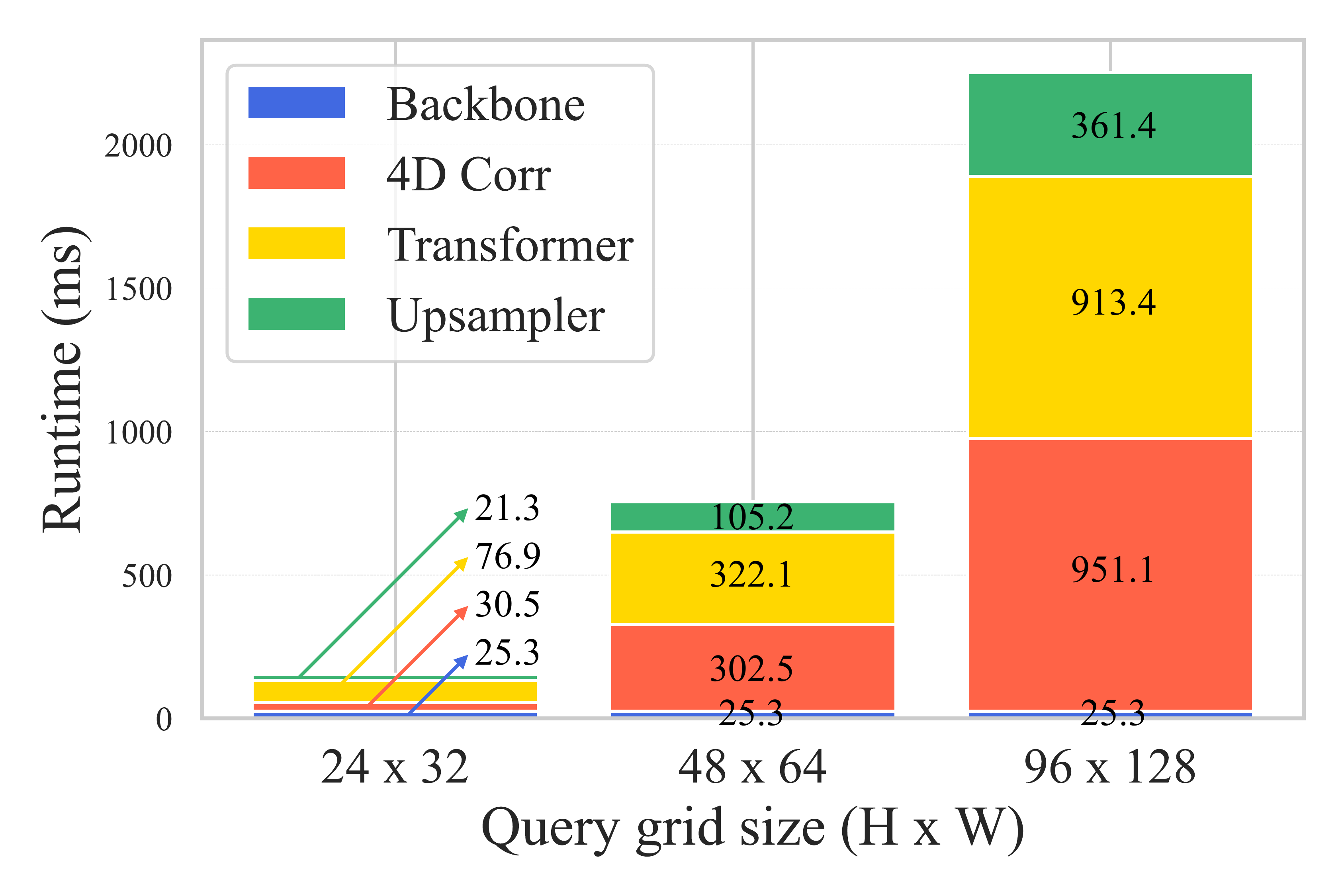}
    \vspace{-7mm}
    \captionof{figure}{Runtime breakdown of DELTA~\cite{ngo2024delta} with a single iteration and one sliding window.}
    \vspace{-5mm}
    \label{fig:component_runtime}
\end{minipage}
\hfill
\begin{minipage}{0.54\linewidth}
    \centering
    \small
    \setlength{\tabcolsep}{1pt}
    \captionof{table}{Comparison of cost reduction strategies on Kubric3D \textit{val} set~\cite{ngo2024delta}. All methods use 1 iteration unless noted; full-resolution outputs are obtained via bilinear interpolation.}
    \begin{tabular}{lcc}
    \toprule
    \textbf{Strategy} & APD$_{3D}\uparrow$ & Runtime (ms)$\downarrow$ \\
    \midrule
    {\color{gray} DELTA (4 iterations)} & \textcolor{gray}{87.3} & \textcolor{gray}{8404} \\
    DELTA (1 iteration) & 74.3 & 2275 \\
    Downsample video reso. ($4\times$) & 35.8 & \cellcolor{custom_green}{\textbf{383}} \\
    Subsample video frames ($4\times$) & \cellcolor{custom_green!30}{65.4} & 833 \\
    Subsample trajectories ($16\times$) & \cellcolor{custom_green}{\textbf{71.2}} & \cellcolor{custom_green!30}{585} \\
    \bottomrule
    \end{tabular}
    
    \label{tab:runtime_strategy}
\end{minipage}
\end{figure*}

Second, through a runtime breakdown for a single iteration of DELTA (see Fig. \ref{fig:component_runtime}), we observe that 4D correlation feature computation becomes increasingly expensive as the number of tracking points grows, contributing to significant runtime overhead. We identify an inefficiency in the implementation used by prior work that leads to suboptimal GPU utilization. To address this, we introduce an effective optimization that significantly reduces runtime without sacrificing performance.

Together, these optimizations achieve a $5\times$ speed-up over DELTA on dense 3D tracking of 100-frame videos, while maintaining state-of-the-art accuracy (Fig.~\ref{fig:teaser}).

In summary, our main contributions are:
1)
a coarse‑to‑fine tracking algorithm that starts with sparsely sampled trajectories and densifies them over iterations;
2)
a learnable interpolator that produces dense trajectories in each iteration and supports adaptive resampling;
3)
an accelerated 4D correlation computation, bringing dense 3D tracking closer to real‑time.

\section{Related Work}
\label{sec:related_work}

\myheading{Optical Flow} estimates motion by establishing dense correspondences between consecutive frames.
Classical variational methods~\cite{memin1998dense, horn1980determining, brox2004high} faced limitations in handling complex dynamics such as fast motion, occlusions, and large displacements.
The advent of CNN-based techniques~\citep{flownet, ranjan2017optical, xu2017accurate, pwcnet} brought substantial improvements to short-term motion estimation.
RAFT~\citep{raft} introduced a key innovation by computing dense correlation volumes across all pixel pairs.
Later approaches expanded on this architecture by introducing transformer-based tokenization of correlation volumes~\citep{flowformer}, employing global feature aggregation to better address occlusions~\citep{jiang2021learning}, and modeling optical flow as a softmax-based matching task~\citep{gmflow}.
Attempts to extend optical flow to long-term sequences using multi-frame methods~\citep{raft,godet2021starflow,shi2023videoflow} or point-tracking integration~\citep{dot,flowtrack} often face challenges with drift and occlusion, limiting long-term tracking performance.



\myheading{Scene Flow} extends optical flow into 3D by estimating dense motion in 3D space. Some methods rely on RGB-D inputs~\citep{hadfield2011kinecting, hornacek2014sphereflow, quiroga2014dense, raft3d, yang2021learning}, while others operate on monocular video alone~\citep{hur2020self, hur2021self, bayramli2023raft, jiang2023emr}. A separate line of work estimates 3D motion directly from point clouds~\citep{flownet3d, flownet3d++, hplflownet, niemeyer2019occupancy}. Several approaches~\citep{he2022learning, liu2019meteornet, huang2022dynamic} leverage temporal context to predict scene flow from videos. More recently, methods such as~\citep{sucar2025dynamic, liang2025zero} build on top of foundation 3D reconstruction models \citep{dust3r, monst3r, mast3r_arxiv24} to estimate 3D scene flow in a more general and scalable manner.

\myheading{Point Tracking} estimates long-range motion trajectories in videos. 
Particle Video~\citep{particle} introduced particle trajectories for long-range video motion. 
TAP-Vid~\citep{tapvid} provided a comprehensive benchmark to evaluate point tracking and TAPNet, a baseline that predicts tracking locations using global cost volume and soft-argmax operation.
PIPs \citep{pips} revisited the concept of particle video and proposed a network that updates trajectories iteratively over fixed temporal windows, but ignored spatial context with independent point tracking and struggle with occlusion. TAPIR \citep{tapir} combines the global estimation of TAPNet \citep{tapvid} and the iterative refinement of \citep{pips} with variable-length window. Subsequent efforts further improve tracking performance by jointly process multiple points with temporal attention and spatial attention~\citep{cotracker}; using 4D correlation feature \citep{cho2024local}. CoTracker3~\citep{cotracker3} proposes a lightweight architecture and a semi-supervised training strategy to further improve the 2D tracking performance. SceneTracker~\citep{scenetracker} and SpaTracker~\citep{spatialtracker} extend point tracking to 3D by incorporating depth information, but remain inefficient for dense tracking due to computationally expensive cross-track attention.
 DELTA \citep{ngo2024delta} is the first approach tackling the dense 3D tracking problem with an efficient transformer and upsample layer to obtain high-resolution dense tracking. 
\textbf{Concurrently}, several other efforts also explore 3D tracking. Seurat~\citep{seurat} estimates depth changes on top of a 2D tracker~\citep{cotracker, cho2024local} to recover 3D motion. TAPIP3D~\citep{tapip3d} proposes a 3D-space correlation mechanism for directly tracking points in 3D space. St4RTrack~\citep{st4rtrack} leverages recent 3D reconstruction models~\citep{dust3r, monst3r} and extends them for dense 3D tracking via joint optimization.

\myheading{Tracking by Reconstructing} tackles long-range motion estimation by explicitly reconstructing a deformable scene representation. OmniMotion~\citep{wang2023tracking} jointly optimizes a NeRF~\citep{mildenhall2020nerf} with a bijective deformation field~\citep{Dinh2016DensityEU} and extracts 2D point trajectories through this mapping. More recent efforts leverage DINOv2~\citep{oquab2023dinov2} to compute long-range correspondences, either through enhanced invertible deformation fields~\citep{Song2024TrackEE} or self-supervised techniques~\citep{dino_tracker_2024}. Shape-of-Motion~\citep{wang2024shape} uses 3D Gaussian Splatting to jointly learn geometry and motion, enabling point tracking by tracing Gaussian positions across frames. While these reconstruction-based methods are capable of producing dense trajectories, they rely on computationally intensive per-video optimization. As a result, they remain significantly slower and less accurate than feedforward, data-driven tracking models on standard benchmarks.
\section{Method}
\label{sec:method}

\begin{figure}[t]
\begin{center}
\includegraphics[width=0.98\linewidth]{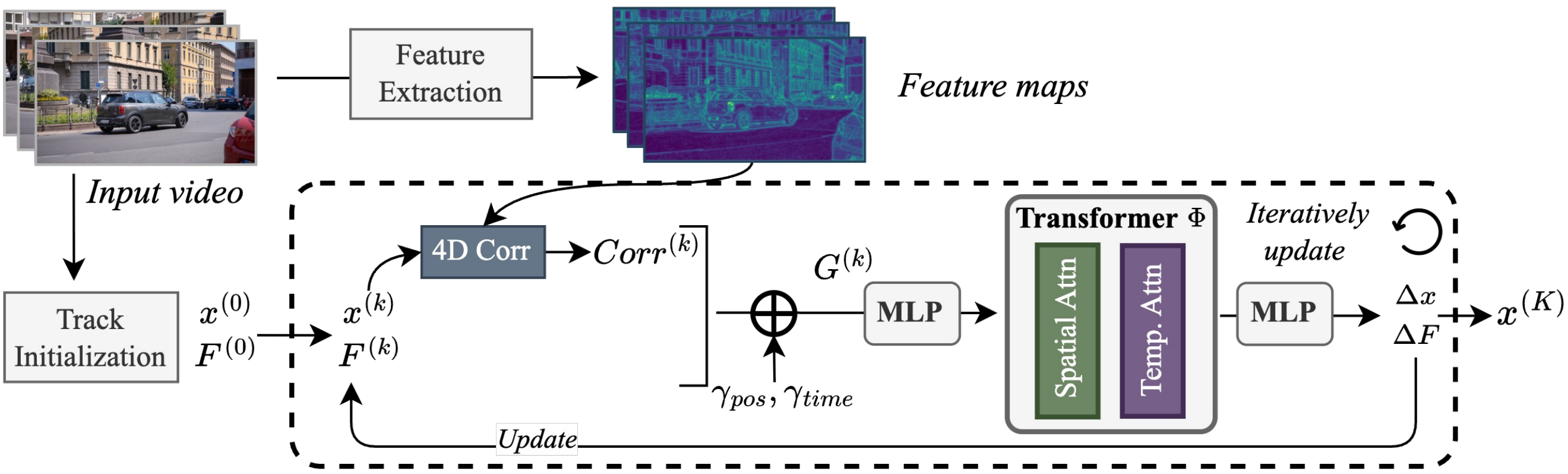}
\captionof{figure}{A modern long-range tracking pipeline (we omit the 3D tracking parts here for simplicity)} 
\label{fig:pipeline}
\end{center}
\vspace{-5mm}
\end{figure}




\myheading{Problem setup:} The input to our method is an RGB-D video, where the RGB frames are represented as \( V \in \mathbb{R}^{T \times H \times W \times 3} \), with \(T\), \(H\), and \(W\) denoting the temporal and spatial dimensions.  
The corresponding depth maps \( D \in \mathbb{R}^{T \times H \times W} \) are generated using a monocular depth estimation model.  
Our model outputs dense, occlusion-aware 3D trajectories \( P \in \mathbb{R}^{T \times H \times W \times 4} \), where each vector  
\( \mathbf{p}_{(t,u,v)} = (u_t, v_t, d_t, o_t) \) captures the motion of a pixel located at \((u, v)\) in the first frame as it propagates to frame \(t\).  
Here, \((u_t, v_t)\) denote the projected 2D pixel coordinates in frame \(t\), \(d_t\) is the predicted depth, and \(o_t \in \{0, 1\}\) indicates the visibility status.


\begin{figure}[t]
\begin{center}
\includegraphics[width=1\linewidth]{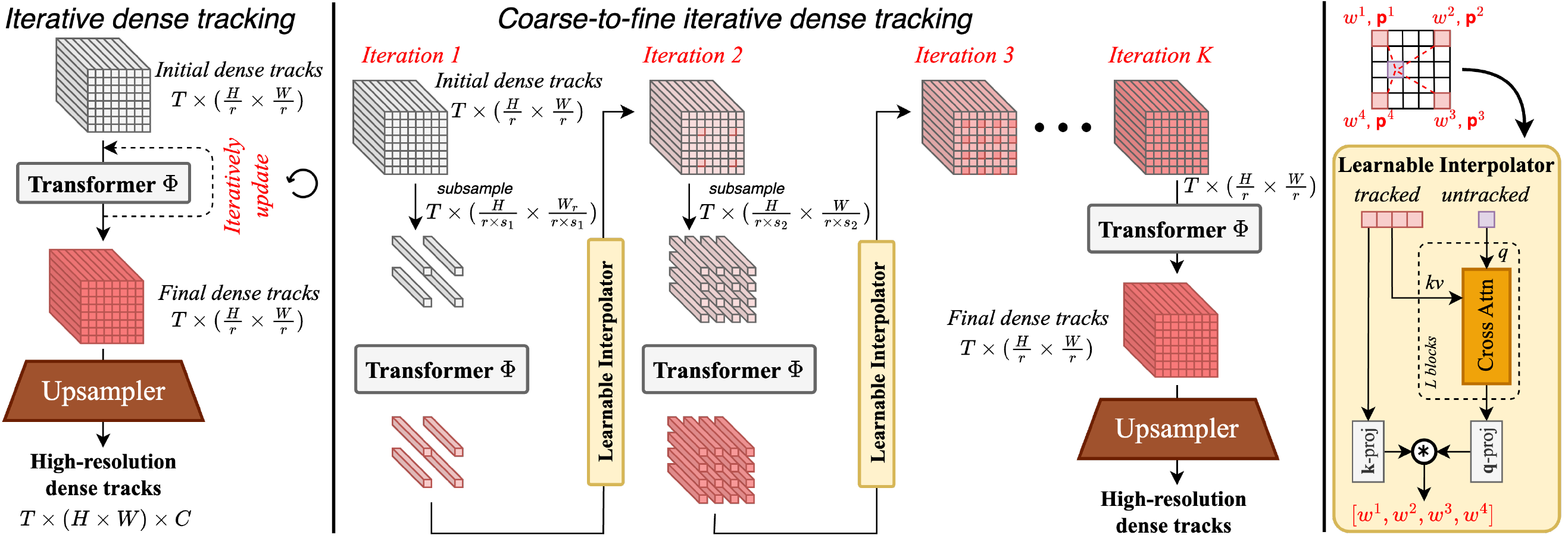}
\captionof{figure}{\textbf{Overview of our proposed framework}. (Left) Traditional iterative dense tracking refines all trajectories at every iteration, leading to high computational cost. (Middle) Our \textbf{coarse-to-fine iterative dense tracking} reduces computation by subsampling trajectory points in early iterations and progressively increasing the density across iterations. (Right) A \textbf{learnable interpolation module} leverages attention to infer untracked motions from nearby tracked pixels, enabling efficient and adaptive trajectory propagation.} 
\label{fig:arch}
\end{center}
\vspace{-5mm}
\end{figure}

\subsection{Preliminary: Modern Tracking Frameworks}
Our work builds upon a modern tracking framework. We select DELTA~\citep{ngo2024delta}, a recent approach for dense 3D tracking from RGB-D videos, but other frameworks have a similar structure (See Fig.~\ref{fig:pipeline} for an overview). 
First, a feature extraction network $F$ (similar to \cite{raft}) creates image feature map for each frame.
Then, the core of the algorithm iteratively evolves a data structure of size of dense tracks 
$\mathcal{D} \in \mathbb{R}^{T \times \frac{H}{r} \times \frac{W}{r}}$, denoted as $\{\emP_i\}$, where $i$ is the trajectory index, and $\emP_i = [\vp_1^i, \vp_2^i, \cdots, \vp_T^i]$, with $\vp_t^i = (u_t^i, v_t^i, d_t^i, o_t^i)$ being the 3D location and visibility of the point associated with the $i$-th trajectory at the $t$-th frame. The downsampling factor $r$ is beneficial for faster processing.

At each iteration, a trajectory is represented by a list of tokens $\emG^i = [\emG_1^i, \emG_2^i, \cdots, \emG_T^i]$, each token $\emG_t^i$  encodes position, visibility, appearance and correlation of the trajectory at $t$-th frame: 
\begin{equation}
G_t^i = [\mathcal{F}_t^i, Corr_t^i, DCorr_t^i, o_t^i, \gamma(\vx_t^i - \vx_1^i)] + \gamma_{pos}(\vx_t^i) + \gamma_{time}(t),
\end{equation}
where $\mathcal{F}_t^i$ are image features extracted by $F$, $Corr_t^i$ are correlation features~\citep{raft, pips, cotracker,cho2024local,ngo2024delta}, $DCorr_t^m$ is depth correlation, and $\gamma_{pos}$ and $\gamma_{time}$ are the positional embedding of the input position $\vx_t^i=(u_t^i,v_t^i,d_t^i)$ and time $t$, respectively.  For more details about the individual components, refer to~\cite{ngo2024delta}.

The tokens $\emG_t^i$ are processed by a transformer network $\Phi$. From the output tokens of $\Phi$ we can compute new dense tracks using a pointwise MLP and use the computed point tracks and the output tokens to compute the input tokens for the next iteration. The framework uses $K$ iterations, e.g., $K=4$.
The network $\Phi$ itself employs a joint global-local spatial attention mechanism: global attention over a sparse set of anchor tracks captures long-range motion, while local attention focuses on fine-grained motion within small spatial neighborhoods.

Finally, a transformer-based upsampling module is used to upsample the dense tracks to the full resolution $T \times H \times W \times 4$. This upsampler models each pixel’s trajectory as a soft combination of neighboring trajectories (see ~\cite{ngo2024delta} for details of this architecture).

While DELTA supports end-to-end dense tracking and has already been optimized for runtime, we undertake the challenging task of finding further significant improvements. We will discuss this next.

\subsection{Analysis of Strategies for Accelerating Tracking}
To accelerate the tracking framework, we first profile the time usage of each stage in the tracking algorithm, as shown in Fig.~\ref{fig:component_runtime}. This analysis reveals two primary bottlenecks: (1) the point tracking transformer $\Phi$, and (2) the computation of the correlation features $Corr$. In this section, we focus on analyzing of the transformer $\Phi$.

The computational cost of $\Phi$ scales linearly with the number of trajectories~\cite{cotracker}, and quadratically with the number of frames. Therefore, reducing either the number of trajectories or the number of frames can lead to substantial speedups. We consider three factors that influence these quantities: the spatial resolution of the input RGB images, the temporal resolution of the video, and the number of trajectory points tracked. To understand the impact of these factors on both performance and efficiency, we design a series of tests to evaluate one iteration of tracking. For each setting, we evaluate two aspects: (1) the time required for computation, and (2) the degradation in tracking accuracy, measured by the $APD_{3D}$ metric on a synthetic dataset. The specific tests are described below.

\noindent\textbf{1) Spatial downsampling of input RGB images:}
We reduce the resolution of the input frames by downsampling both the width and height by a factor of 4. Point tracking is then performed on the low-resolution frames using the pretrained transformer $\Phi$. We then upsample the resulting trajectories back to the original resolution for evaluation. This analysis reduces both the number of trajectories and the spatial cost of attention, providing insights into how resolution affects accuracy and speed.

\noindent\textbf{2) Temporal subsampling of input frames:}
We subsample the input video along the temporal dimension by a factor of 4, reducing the number of frames used in trajectory computation. The trajectories are still computed at full spatial resolution. Afterward, we apply linear interpolation to upsample the trajectories back to match the original temporal length of the video. Although more advanced interpolation methods such as B-splines or non-linear motion kernels could be used to model more complex dynamics, linear interpolation is sufficient for analyzing the effects of temporal subsampling on performance.

\noindent\textbf{3) Subsampling the number of trajectories:}
We keep the input frames at full spatial and temporal resolution, but subsample the initial set of trajectory points by a factor of 16. This reduces the number of tokens passed through the transformer while preserving the original feature resolution. 

{4) \bf Baseline:} no downsampling is used as reference. We denote this as DELTA (1 iteration) as reference and DELTA (4 iterations) for the complete algorithm.

We present the results of this analysis in Tab.~\ref{tab:runtime_strategy} (see the \textit{supplementary} for more details). Among the evaluated strategies, \textit{subsampling the number of trajectories} emerges as the most effective, as other approaches incur significantly higher tracking errors with clear drawbacks. \textit{Spatial downsampling} fails due to information loss caused by reduced feature map resolution.  \textit{Temporal subsampling} assumes linear motion over time, but DELTA tracks in UVD space, where 3D linear motion does not correspond to linear 2D and depth trajectories. As a result, interpolating skipped frames introduces distortion and lowers accuracy. Based on this analysis, we propose a new algorithm that leverages trajectory subsampling as a core component (see Sec.~\ref{sec:coarse-to-fine}). To complement this, we introduce a novel learnable interpolation architecture to recover dense trajectories, forming a key component of our method (see Sec.~\ref{sec:interpolator}). Finally, we present an accelerated design for computing correlation features, further improving overall efficiency (see Sec.~\ref{sec:correlation}).

In addition to the three major contributions discussed above, we explored several alternative design choices to further accelerate the tracking framework. These include reducing the number of transformer layers (e.g., from 6 to 3, see Tab.~\ref{tab:ablation_runtime}), lowering the number of iterative refinements (see Fig.~\ref{fig:runtime_iterations}), and experimenting with various trajectory subsampling strategies (see Tab.~\ref{tab:ablation_sampling}).

\subsection{Accelerating Dense Point Tracking}

%


\subsubsection{Coarse to Fine Tracking with Trajectory Subsampling}
\label{sec:coarse-to-fine}
As shown in Tab.~\ref{tab:runtime_strategy}, tracking on a subsampled set of points followed by interpolation significantly reduces runtime while maintaining accuracy for the interpolated points. This observation naturally leads to a coarse-to-fine tracking algorithm. We begin by sampling a sparse set of points and perform one iteration of point tracking. Next, we interpolate the positions of the remaining untracked points. These interpolated positions provide reasonable estimates that serve as initialization for a denser set of points in the next iteration. The process is repeated, using the interpolated estimates to guide the position updates for increasingly dense sets of points, progressively refining the tracking results.

More concretely, the algorithm proceeds as follows.

\textbf{1) Subsample points.} We begin by subsampling a small set of points to track from a $\frac{H}{r} \times \frac{W}{r}$ pixel grid, where $r$ is the upsampling ratio of the upsampler used in DELTA. In our implementation (illustrated in Fig.~\ref{fig:arch}), we uniformly sample points on a sparse grid of size $\frac{H}{r \times s_1} \times \frac{W}{r \times s_1}$, where $s_1$ denotes the initial spatial subsampling scale for the first iteration.

Each sampled point is initialized with its original pixel position. After each coarse-to-fine tracking iteration, we update the positions of the points that have already been tracked and introduce additional points for tracking in the next iteration. At the $k$-th iteration, the number of points tracked corresponds to a grid of size $\frac{H}{r \times s_k} \times \frac{W}{r \times s_k}$, where $s_k$ is the subsampling scale at step $k$. In our implementation, we decrease $s_k$ by a factor of 2 at each step, progressively refining the tracking resolution.

In our ablation study, we compare different subsampling schedules and explore alternative strategies, such as random sampling and importance sampling focused on distinctive features (e.g., object boundaries). However, we find that the simple uniform grid sampling performs on par with these more complex strategies. Therefore, we adopt the grid-based approach for its simplicity.

\noindent\textbf{2) Single-step point tracking.} We compute the correlation features for each sampled point based on its current estimated or initialized position, and feed these features into the transformer $\Phi$ to produce updated point positions.

\noindent\textbf{3) Interpolate positions for new points.}
Given the current estimated positions of the sampled points, we use an interpolation method to estimate the initial positions of the additional points introduced at the next scale, which have not yet been tracked. We experimented with various interpolation strategies, including bilinear and nearest-neighbor interpolation, and found that nearest-neighbor consistently outperforms bilinear (see Fig.~\ref{fig:runtime_iterations}). We hypothesize that bilinear interpolation tends to oversmooth positional estimates, especially in regions with large motion differences, which is a problem that becomes more pronounced in long-term tracking scenarios. To further improve the accuracy of interpolated positions, we introduce a learnable interpolation module (see Sec.~\ref{sec:interpolator}), which is trained end-to-end alongside the coarse-to-fine tracking pipeline.

The algorithm then returns to step \textbf{1)} and repeats the process until the full resolution is reached.

\subsubsection{Learnable Interpolation Module}
\label{sec:interpolator}
To propagate motion from tracked to untracked pixels during coarse-to-fine iterations, we introduce a \textit{learnable interpolation module} that estimates the 3D motion of each untracked pixel as an adaptive blend of nearby tracked motions. Unlike fixed interpolation methods (e.g., bilinear or nearest-neighbor), our approach dynamically predicts interpolation weights using attention over spatial features. Let $\mathcal{P}_{\text{track}}$ be the set of tracked pixels and $\mathcal{P}_{\text{query}}$ the untracked ones. For each query $(u,v) \in \mathcal{P}_{\text{query}}$, we predefine its four nearest neighbors $\{(u_j, v_j)\}_{j=1}^{4} \in \mathcal{P}_{\text{track}}$. The interpolated 3D motion $\mathbf{p}_{(u,v)}$ (for clarity, we omit the frame subscript $t$) is computed as:
\vspace{-2mm}
\begin{equation}
\vspace{-2mm}
\mathbf{p}_{(u,v)} = \sum_{j=1}^{4} w^j_{(u,v)} \cdot \mathbf{p}_{(u_j, v_j)}
\label{eq:interp-motion}
\end{equation}
where weights $w^j_{(u,v)} \in [0,1]$ sum to 1 and are predicted by a lightweight attention module.

Let $\mathcal{F} \in \mathbb{R}^{\frac{H}{r} \times \frac{W}{r} \times C_{\mathcal{F}}}$ denote the dense feature map of the query frame of the dense tracking from the feature extractor $F$. The initial query feature $\mathcal{F}_{(u,v)}$ is first refined via $L$ multi-head cross-attention blocks over the support set. We follow the Alibi attention scheme~\citep{alibi} in these blocks, where a relative positional bias is added to the attention logits, computed as the L1 distance between the query and support pixel locations. The refined query and the final weights are predicted as:
\begin{equation}
\tilde{\mathcal{F}}_{(u,v)} = \text{CrossAttn}^{(L)}\left( \mathcal{F}_{(u,v)}, \left\{\mathcal{F}_{(u_j, v_j)}\right\} \right), \:
\mathbf{w}_{(u,v)} = \text{softmax} \left( \vq(\tilde{\mathcal{F}}_{(u,v)}) \cdot \vk\left(\left\{\mathcal{F}_{(u_j, v_j)}\right\}\right) \right)
\label{eq:attn-refine-and-weights}
\end{equation}


where $\vq(\cdot)$ and $\vk(\cdot)$ are linear projections for the refined query and support features. This mechanism allows sparse trajectories to efficiently guide the evolution of the full dense motion field over time, maintaining temporal/spatial consistency without the cost of processing full dense trajectories.

\subsubsection{Accelerating Computation of Correlation Features}
\label{sec:correlation}

A major bottleneck in DELTA is the computation of 4D correlation features, introduced in~\cite{cho2024local}, which measure pairwise similarities between local neighborhoods around query and predicted positions. These features form a 4D tensor processed by a dual-convolutional module. As shown in our profiling (Fig.~\ref{fig:component_runtime}), this becomes increasingly expensive with a large number of trajectories.

The main inefficiency arises from the input channel size (typically $7 \times 7 = 49$), which is not divisible by 8, leading to poor GPU utilization in dense settings. To address this, we add a lightweight MLP projection that reduces the dimension to 32, followed by LayerNorm and ReLU. This adds minimal overhead for sparse tracking and yields a significant speedup in the dense setting. An alternative from~\cite{cotracker3} replaces the dual-conv module with pure MLPs on flattened 4D correlation, but we find it performs worse on dense 2D/3D tracking (See Tab.~\ref{tab:ablation_runtime}).

\section{Experiments}
\label{sec:exp}

\begin{table}
\centering
\small
\caption{\textbf{Long-range optical flow results} on CVO  \citep{accflow, dot}.}
\setlength{\tabcolsep}{4pt}
\resizebox{0.85\linewidth}{!}{
\footnotesize
\begin{tabular}{lcccccc}
\toprule
\multirow{2}{*}{\textbf{Methods}} & \multicolumn{2}{c}{\textbf{CVO-Clean} (7 frames)} & \multicolumn{2}{c}{\textbf{CVO-Final} (7 frames)} & \multicolumn{2}{c}{\textbf{CVO-Extend} (48 frames)} \\
 & EPE$\downarrow$ (\textit{all/vis/occ}) & IoU$\uparrow$ & EPE $\downarrow$ (\textit{all/vis/occ}) & IoU$\uparrow$ & EPE$\downarrow$ (\textit{all/vis/occ}) & IoU$\uparrow$ \\
\midrule
RAFT &  2.48 / 1.40 / 7.42 & 57.6 & 2.63 / 1.57 / 7.50 & 56.7 & 21.80 / 15.4 / 33.4 & 65.0 \\
MFT &  2.91 / 1.39 / 9.93 & 19.4 & 3.16 / 1.56 / 10.3 & 19.5 & 21.40 / 9.20 / 41.8 & 37.6 \\ \cdashline{1-7}
TAPIR & 3.80 / 1.49 / 14.7 & 73.5 & 4.19 / 1.86 / 15.3 & 72.4 & 19.8 / 4.74 / 42.5 & 68.4 \\
CoTracker2 & 1.51 / 0.88 / 4.57 & 75.5 & 1.52 / 0.93 / 4.38 & 75.3 & 5.20 / 3.84 / 7.70 & 70.4  \\
DOT & 1.29 / 0.72 / 4.03 & \cellcolor{custom_green}\textbf{80.4} & 1.34 / 0.80 / 3.99 & \cellcolor{custom_green}\textbf{80.4} & 4.98 / 3.59 / 7.17 & \cellcolor{custom_green}\textbf{71.1} \\ \cdashline{1-7}
SceneTracker & 4.40 / 3.44 / 9.47 & - & 4.61 / 3.70 / 9.62 & - & 11.5 / 8.49 / 17.0 & -   \\
SpatialTracker & 1.84 / 1.32 / 4.72 & 68.5  & 1.88 / 1.37 / 4.68 & 68.1 & 5.53 / 4.18 / 8.68 & 66.6  \\
DOT-3D & 1.33 / 0.75 / 4.16 & \cellcolor{custom_green!30}{79.0} & 1.38 / 0.83 / 4.10 & \cellcolor{custom_green!30}{78.8} & 5.20 / 3.58 / 7.95 & \cellcolor{custom_green!30}{70.9} \\
DELTA  &\cellcolor{custom_green}\textbf{ 0.94} / \cellcolor{custom_green}\textbf{0.51} / \cellcolor{custom_green}\textbf{2.97} & 78.7 & \cellcolor{custom_green}{\textbf{1.03} / \textbf{0.61} / \textbf{3.03}} & 78.3 & 3.67 / 2.64 / 5.30 & 70.1 \\
\cdashline{1-7}
Ours  & \cellcolor{custom_green!30}{1.04 / 0.61 / 3.25} & 77.6 & \cellcolor{custom_green!30}{1.12 / 0.69 / 3.27} & 77.3 & \cellcolor{custom_green}\textbf{3.53} / \cellcolor{custom_green}\textbf{2.57} / \cellcolor{custom_green}\textbf{5.10} & 70.6  \\
\bottomrule
\end{tabular}
}

\label{tab:2d_dense_tracking_results}
\end{table}




\myheading{Implementation Details:}
Following prior work~\citep{ngo2024delta, cotracker, spatialtracker}, we use the Kubric engine~\citep{kubric} to generate 5,631 RGB-D training videos with both sparse and dense tracking annotations. During training, we apply online data augmentation, including color jittering, random scaling, and random cropping.
As in~\cite{ngo2024delta}, we supervise the model on both sparse and dense tracks. The total loss combines the 2D coordinate loss $\mathcal{L}_{2D}$, inverse depth loss $\mathcal{L}_{depth}$, and visibility loss $\mathcal{L}_{visib}$. The loss weights $\lambda_{2d}, \lambda_{depth}, \lambda_{visib}$ follow those in~\cite{ngo2024delta}.
We initialize the model from the pretrained DELTA checkpoint and add parameters for the interpolation module. The model is trained for 100{,}000 iterations on 8 A100 GPUs using AdamW with a one-cycle learning rate schedule, starting from an initial learning rate of $2 \times 10^{-4}$.


\subsection{Comparison with prior work}

\noindent\textbf{Baselines.} 
We compare our method with previous optical flow and point tracking approaches. In particular, we closely compare it with DELTA~\cite{ngo2024delta}, the current SOTA for dense 3D tracking. We also include DOT~\cite{dot}, a method for dense 2D tracking, its 3D variant DOT-3D \cite{ngo2024delta}, and other recent point tracking methods, including CoTracker2~\cite{cotracker}, CoTracker3~\cite{cotracker3}, SpaTracker~\cite{spatialtracker}, and SceneTracker~\cite{scenetracker}.

\myheading{Benchmark Datasets.}  
We evaluate our method on a diverse set of tracking benchmarks covering 2D optical flow, dense 3D pixel tracking, and sparse 3D point tracking. (1) Long-range 2D optical flow: We use the \textbf{CVO} dataset~\citep{accflow, dot}, which includes three splits: \textit{CVO-Clean}, \textit{CVO-Final}, and \emph{CVO-Extended}. Each split contains about 500 videos with 7 frames (48 frames for \emph{CVO-Extended}), captured at 60 FPS, with dense long-range optical flow annotations and occlusion masks. (2) Dense 3D pixel tracking: Following~\cite{ngo2024delta}, we evaluate on a held-out test split of the \textbf{Kubric} dataset~\citep{kubric}, which includes 143 RGB-D videos, each 24 frames long, with dense ground-truth 3D trajectories for every pixel. (3) 3D point tracking: We benchmark on the large-scale \textbf{TAP-Vid3D} dataset~\citep{tapvid3d}, which contains 4,569 videos from DriveTrack~\citep{balasingam2024drivetrack}, PStudio~\citep{joo2017panoptic}, and Aria~\citep{pan2023aria}, covering a range of real-world and simulated environments, with video lengths from 25 to 300 frames.





\myheading{Metrics.} For the \emph{long-range optical flow}, we adopt the evaluation protocol from~\cite{dot,ngo2024delta}, reporting the end-point error (EPE) between predicted and ground-truth flows and the occlusion prediction accuracy using the intersection-over-union (IoU) between the predicted and ground-truth visibility masks. For the \emph{dense 3D tracking} and \emph{sparse 3D point tracking} benchmarks, we follow the metrics proposed in~\cite{tapvid3d}. Specifically, we report APD$_{3D}$ (Average Percent of Points within a threshold to measure spatial accuracy, OA (Occlusion Accuracy) for evaluating visibility prediction, and AJ (Average Jaccard), which jointly captures both spatial and occlusion correctness.

\myheading{Long-range 2D optical flow.} Our method achieves comparable performance to \textsc{DELTA}~\cite{ngo2024delta}, a state-of-the-art approach for dense tracking (see Tab.~\ref{tab:2d_dense_tracking_results}). While we observe slightly lower accuracy on the \textit{Clean} and \textit{Final} subsets, our model outperforms \textsc{DELTA} on the more challenging \textit{Extended} subset, which contains significantly longer video sequences. This demonstrates our model's robustness in long-range tracking scenarios. 
\vspace{-3mm}

\myheading{Dense 3D Tracking.} Table~\ref{tab:3d_dense_tracking} reports results on the Kubric3D test set. Our method matches the performance of DELTA while significantly outperforming other baselines, including 3D tracking methods such as SpaTracker\cite{spatialtracker} and SceneTracker\cite{scenetracker}, as well as 2D-to-3D lifted approaches like CoTracker2\cite{cotracker} and CoTracker3\cite{cotracker3}. Importantly, our approach is substantially more efficient: it achieves up to a $5\times$ speed-up over \textsc{DELTA} and is approximately $100\times$ faster than \textsc{SpaTracker}, enabling practical deployment in real-time or large-scale applications. Qualitative results of dense 3D tracking on in-the-wild videos are provided in the s\textit{upplementary materials}.
\vspace{-3mm}
\begin{table}
\centering
\small
\caption{\textbf{Dense 3D tracking results} on the Kubric3D dataset.}
\footnotesize
\begin{tabular}{lcccc}
\toprule
\multirow{2}{*}{\textbf{Methods}} & \multicolumn{3}{c}{\textbf{Kubric-3D} (24 frames)} & \textbf{Runtime}$\downarrow$ \\
 & AJ$\uparrow$ (\textit{all/vis}) & APD$_{3D}\uparrow$ (\textit{all/vis}) & OA$\uparrow$ & (second) \\
\midrule
CoTracker2 & 70.0 / 80.7 & 76.4 / 85.1 & 96.7 & 145  \\
CoTracker3 & 68.9 / 79.3 & 75.9 / 84.3 & 95.8 & 94  \\
\addlinespace[1pt]
\cdashline{1-5}
\addlinespace[1pt]
SpatialTracker & 35.3 / 42.7 & 49.1 / 51.6 & 96.5 &  350 \\
SceneTracker & - & 45.0 / 65.5 & - &  179\\
DOT-3D & 68.1 / 72.3 & 75.3 / 77.5 & 88.7 &  \cellcolor{custom_green!30}11.8 \\
DELTA & \textbf{81.7} / \cellcolor{custom_green}\textbf{85.1} & \cellcolor{custom_green!30}87.5 / 90.9 & \cellcolor{custom_green}\textbf{97.5}& 18.2 \\
\cdashline{1-5}
Ours & \cellcolor{custom_green}{\textbf{81.7}} / \cellcolor{custom_green!30}{84.8} & \cellcolor{custom_green}\textbf{87.6} / \cellcolor{custom_green}\textbf{91.0} & \cellcolor{custom_green!30}97.0 & \cellcolor{custom_green}\textbf{3.5} \\
\bottomrule
\end{tabular}

\label{tab:3d_dense_tracking}
\vspace{-2mm}
\end{table}

\begin{table}
\centering
\caption{\textbf{3D tracking results} on the TAP-Vid3D Benchmark. We use UniDepth for depth estimation.$^{\dagger}$ denotes using depth to lift 2D tracks to 3D tracks.}
\setlength{\tabcolsep}{1pt}
\resizebox{\linewidth}{!}{
\footnotesize
\begin{tabular}{lccc|ccc|ccc|ccc}
\toprule
\multirow{2}{*}{\textbf{Methods}} & \multicolumn{3}{c|}{\textbf{Aria}} & \multicolumn{3}{c|}{\textbf{DriveTrack}} & \multicolumn{3}{c|}{\textbf{PStudio}} & \multicolumn{3}{c}{\textbf{Average}} \\
 & AJ$\uparrow$ & APD$_{3D}\uparrow$ & OA$\uparrow$ & AJ$\uparrow$ & APD$_{3D}\uparrow$ & OA$\uparrow$ & AJ$\uparrow$ & APD$_{3D}\uparrow$ & OA$\uparrow$ & AJ$\uparrow$ & APD$_{3D}\uparrow$& OA$\uparrow$ \\
\midrule
TAPIR$^{\dagger}$ + COLMAP & 7.1 & 11.9 & 72.6 & 8.9 & 14.7 & 80.4 & 6.1 & 10.7 & 75.2 & 7.4 & 12.4 & 76.1 \\
CoTracker2$^{\dagger}$ + COLMAP & 8.0 & 12.3 & 78.6 & 11.7 & 19.1 & 81.7 & 8.1 & 13.5 & 77.2 & 9.3 & 15.0 & 79.1 \\
BootsTAPIR$^{\dagger}$ + COLMAP & 9.1 & 14.5 & 78.6 & 11.8 & 18.6 & 83.8 & 6.9 & 11.6 & \cellcolor{custom_green!30}{81.8} & 9.3 & 14.9 & 81.4 \\
\cdashline{1-13}
CoTracker2$^{\dagger}$ + UniDepth & 13.0 & 20.9 & 84.9 & 12.5 & 19.9 & 80.1 & 6.2 & 13.5 & 67.8 & 10.6 & 18.1 & 77.6 \\
TAPTR$^{\dagger}$ + UniDepth & 15.7 & 24.2 & \cellcolor{custom_green!30}{87.8} & 12.4 & 19.1 & \cellcolor{custom_green!30}{84.8} & 7.3 & 13.5 & \cellcolor{custom_green}{\textbf{84.3}} & 11.8 & 18.9 & \cellcolor{custom_green}\textbf{85.6} \\
LocoTrack$^{\dagger}$ + UniDepth & 15.1 & 24.0 & 83.5 & 13.0 & 19.8 & 82.8 & 7.2 & 13.1 & 80.1 & 11.8 & 19.0 & 82.3 \\
\cdashline{1-13}
SpatialTracker + UniDepth & 13.6 & 20.9 & \cellcolor{custom_green}\textbf{90.5} & 8.3 & 14.5 & 82.8 & 8.0 & \cellcolor{custom_green}\textbf{15.0} & 75.8 & 10.0 & 16.8 & \cellcolor{custom_green!30}{83.0} \\
SceneTracker + UniDepth & - & 23.1 & - & - & 6.8 & - & - & 12.7 & - & - & 14.2 & - \\
DOT-3D + UniDepth & 13.8 & 22.1 & 85.5 & 11.8 & 17.9 & 82.3 & 3.2 & 5.3 & 52.5 & 9.6 & 15.1 & 73.4 \\
DELTA + UniDepth & \cellcolor{custom_green!30}{16.6} & \cellcolor{custom_green!30}{24.4} & 86.8 & \cellcolor{custom_green!30}{14.6} & \cellcolor{custom_green!30}{22.5} & \cellcolor{custom_green}\textbf{85.8} & \cellcolor{custom_green}\textbf{8.2} & \cellcolor{custom_green}\textbf{15.0} & 76.4 & \cellcolor{custom_green!30}{13.1} & \cellcolor{custom_green!30}{20.6} & \cellcolor{custom_green!30}{83.0} \\
\cdashline{1-13}
Ours + UniDepth & \cellcolor{custom_green}\textbf{17.0} & \cellcolor{custom_green}\textbf{24.7} & 87.2 & \cellcolor{custom_green}\textbf{15.6} & \cellcolor{custom_green}\textbf{23.8} & 84.6 & \cellcolor{custom_green!30}{7.7} & \cellcolor{custom_green!30}{14.4} & 74.1 & \cellcolor{custom_green}\textbf{13.4} & \cellcolor{custom_green}\textbf{21.0} & 81.2 \\
\bottomrule

\end{tabular}
}
\label{tab:3d_tracking_results_combined}
\vspace{-5mm}
\end{table}


\myheading{3D Tracking.} We evaluate our approach on sparse 3D point tracking in Table~\ref{tab:3d_tracking_results_combined}. Although the sparse setting does not directly benefit from our coarse-to-fine strategy, the results demonstrate that training with this strategy—and incorporating our redesigned 4D correlation module—does not degrade sparse tracking performance. Our method achieves slightly better results on the \textit{Aria} and \textit{DriveTrack} subsets, with a minor drop on \textit{PStudio}, showing overall competitive and robust performance.



\begin{figure}[t]
\begin{center}
\includegraphics[width=\linewidth]{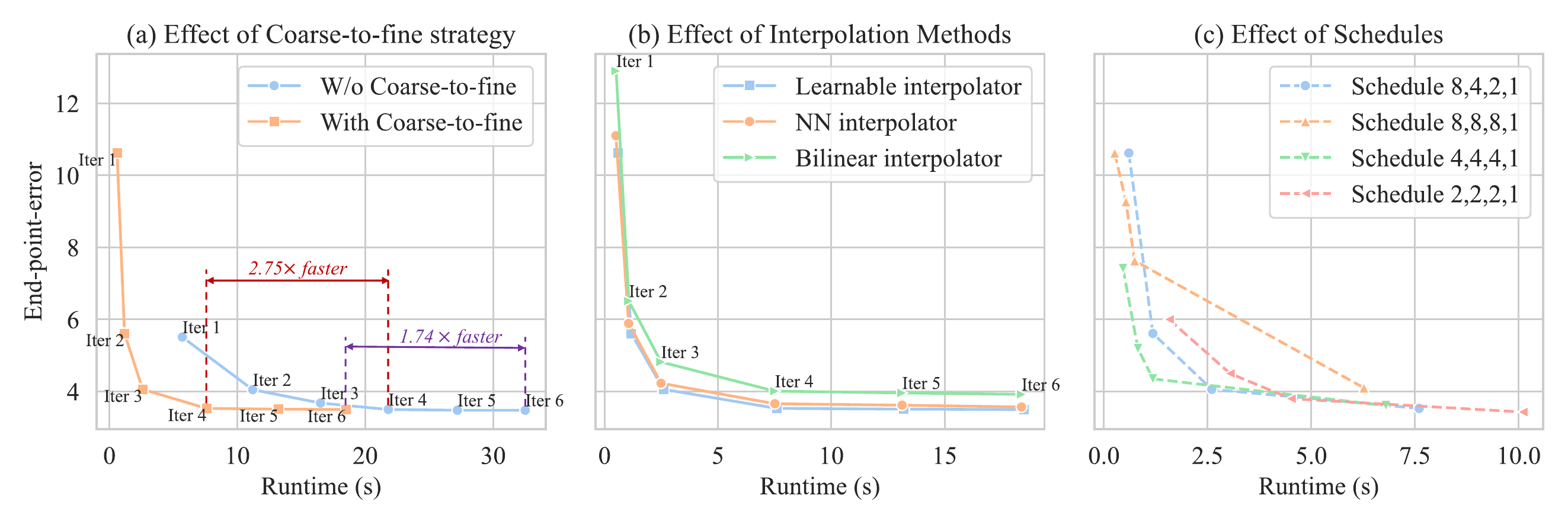}
\vspace{-6mm}
\captionof{figure}{\textbf{Analysis of the coarse-to-fine strategy.}  We visualize how accuracy evolves with runtime across different methods. (a) We evaluate runtime per iteration and observe that the coarse-to-fine strategy consistently reduces runtime compared to the baseline while achieving similar accuracy. (b) Nearest neighbor interpolation outperforms bilinear, and our learnable interpolation further improves accuracy. (c) We compare different coarse-to-fine scheduling strategies.}
\label{fig:runtime_iterations}
\end{center}
\vspace{-4mm}
\end{figure}


\subsection{Ablation Study}



     

\begin{table}[t]
\centering
\small

\begin{minipage}[t]{0.48\linewidth}
    \centering
    \captionof{table}{Ablation of coarse-to-fine sampling strategies.}
    \label{tab:ablation_sampling}
    \footnotesize
    \setlength{\tabcolsep}{1pt}
    \begin{tabular}{lcc}
    \toprule
    \textbf{Sampling} & EPE$\downarrow$ & Runtime (s) \\
    \midrule
    None & 3.50 / 2.55 / 5.09 & 21.8 \\
    Uniform grid {\color[RGB]{42,157,143}(default)} & 3.53 / 2.57 / 5.10 & 7.6 \\
    Random & 3.51 / 2.61 / 5.04 & 7.8 \\
    SIFT & 3.79 / 2.88 / 5.72 & 7.8 \\
    \bottomrule
    \end{tabular}
\end{minipage}
\hfill
\begin{minipage}[t]{0.48\linewidth}
    \centering
    \captionof{table}{Impact of runtime optimization components.}
    \label{tab:ablation_runtime}
    \footnotesize
    \begin{tabular}{lcc}
    \toprule
    \textbf{Model Variant} & EPE$\downarrow$ & Runtime (s) \\
    \midrule
    DELTA~\cite{ngo2024delta} & 3.67 / 2.64 / 5.30 & 32.1 \\
    + new 4D Corr & \cellcolor{custom_green}{3.50 / 2.55 / 5.09} & 21.8 \\
    + Coarse-to-fine & \cellcolor{custom_green!20}{3.53 / 2.57 / 5.10} & 7.6 \\
    + 3-layer Trans & \cellcolor{red!20}{3.91 / 2.85 / 5.82} & 5.8 \\
    + MLP 4D Corr & \cellcolor{red!40}{4.76 / 2.81 / 7.75} & 5.4 \\
    \bottomrule
    \end{tabular}
\end{minipage}
\vspace{-5mm}
\end{table}
    
We conduct a series of ablation studies on the CVO \textit{Extended} split~\cite{dot}. Runtime is measured as the average time required to densely track all $384 \times 512$ pixels across a 48-frame video using a machine with a single A100 GPU. The number of iterations is set to 4 by default, unless otherwise specified.

\myheading{Analysis of the Coarse-to-fine strategy.} Figure~\ref{fig:runtime_iterations}a compares our coarse-to-fine strategy (schedule: 8,4,2,1) with the baseline that tracks all trajectories in every iteration. Our method achieves similar accuracy while being $2.75\times$ faster. In Fig.\ref{fig:runtime_iterations}b, we ablate different interpolators, showing that our attention-based design outperforms fixed alternatives like nearest-neighbor and bilinear. Fig.~\ref{fig:runtime_iterations}c compares different subsampling schedules, where $(s_1,s_2,s_3,s_4)$ denotes the subsampling factors across 4 iterations. A finer schedule, such as (2,2,2,1), improves accuracy but with increased runtime.

Table~\ref{tab:ablation_sampling} presents different pixel sampling strategies. In addition to the default uniform grid sampling, we evaluate two alternatives: random sampling and SIFT-based keypoint selection. Both alternatives require an additional step to identify neighbors for the interpolation module. While the random strategy yields slightly better results, the SIFT-based approach performs worse, likely due to non-uniform coverage. We retain the uniform grid sampling as our default due to its simplicity, efficiency, and competitive performance.

\myheading{Runtime Analysis.} Table~\ref{tab:ablation_runtime} summarizes our runtime improvements. Replacing the original 4D correlation in \textsc{DELTA}~\cite{ngo2024delta} reduces runtime by 33\%. Adding our coarse-to-fine strategy yields a further $3\times$ speed-up. A 3-layer transformer brings runtime to 5.8s, though with some loss in accuracy.

\section{Conclusion}
\label{sec:conclusion}



In this work, we presented an efficient framework for dense 3D video tracking that significantly improves the runtime of DELTA \cite{ngo2024delta} while preserving its strong performance. We proposed a coarse-to-fine tracking algorithm that progressively increases spatial coverage across iterations, combined with a learnable interpolation module for dense supervision and a faster 4D correlation computation. These improvements yield $5-100\times$ speedup over previous approaches, making our approach more suitable for real-time applications. Nonetheless, our method inherits common \textbf{limitations} of data-driven tracking: it is trained on synthetic data and may struggle under fast motion, severe occlusions, or poor depth estimation, which affect 2D/3D tracking accuracy in complex real-world scenes.

 \paragraph{Acknowledgements} Evangelos Kalogerakis has received funding from the European Research Council (ERC) under the Horizon research and innovation programme (Grant agreement No. 101124742).

\clearpage
\bibliographystyle{plain}
\bibliography{neurips_2025}

\appendix

\section{More Results}
\subsection{3D Tracking}

We further evaluate our method on the TAP-Vid3D benchmark using depth maps predicted by ZoeDepth~\citep{zoedepth}. Results are summarized in Tab.~\ref{tab:3d_tracking_results_zoedepth}.

\begin{table}
\centering
\caption{\textbf{3D tracking results} on the TAP-Vid3D Benchmark. We use ZoeDepth for depth estimation.$^{\dagger}$ denotes using depth to lift 2D tracks to 3D tracks.}\setlength{\tabcolsep}{1pt}
\footnotesize
\begin{tabular}{lccc|ccc|ccc|ccc}
\toprule
\multirow{2}{*}{\textbf{Methods}} & \multicolumn{3}{c|}{\textbf{Aria}} & \multicolumn{3}{c|}{\textbf{DriveTrack}} & \multicolumn{3}{c|}{\textbf{PStudio}} & \multicolumn{3}{c}{\textbf{Average}} \\
 & AJ$\uparrow$ & APD$_{3D}\uparrow$ & OA$\uparrow$ & AJ$\uparrow$ & APD$_{3D}\uparrow$ & OA$\uparrow$ & AJ$\uparrow$ & APD$_{3D}\uparrow$ & OA$\uparrow$ & AJ$\uparrow$ & APD$_{3D}\uparrow$& OA$\uparrow$ \\
\midrule
TAPIR$^{\dagger}$ + ZoeDepth & 9.0 & 14.3 & 79.7 & 5.2 & 8.8 & 81.6 & 10.7 & 18.2 & 78.7 & 8.3 & 13.8 & 80.0 \\
CoTracker2$^{\dagger}$ + ZoeDepth & \cellcolor{custom_green!30}{10.0} & 15.9 & \cellcolor{custom_green!30}{87.8} & 5.0 & 9.1 & 82.6 & \cellcolor{custom_green!30}{11.2} & \cellcolor{custom_green}{\textbf{19.4}} & 80.0 & 8.7 & 14.8 & 83.4 \\
BootsTAPIR$^{\dagger}$ + ZoeDepth & 9.9 & \cellcolor{custom_green}{16.3} & 86.5 & 5.4 & 9.2 & 85.3 & \cellcolor{custom_green}{\textbf{11.3}} & \cellcolor{custom_green!30}{19.0} & \cellcolor{custom_green!30}{82.7} & 8.8 & 14.8 & \cellcolor{custom_green!30}{84.8} \\
TAPTR$^{\dagger}$ + ZoeDepth & 9.1 & 15.3 & \cellcolor{custom_green!30}{87.8} & 7.4 & 12.4 & 84.8 & 10.0 & 17.8 & \cellcolor{custom_green}{\textbf{84.3}} & 8.8 & 15.2 & \cellcolor{custom_green}{\textbf{85.6}} \\
LocoTrack$^{\dagger}$ + ZoeDepth & 8.9 & 15.1 & 83.5 & 7.5 & 12.3 & 82.8 & 9.7 & 17.1 & 80.1 & 8.7 & 14.8 & 82.1 \\
\cdashline{1-13}
SpatialTracker + ZoeDepth & 9.2 & 15.1 & \cellcolor{custom_green}{\textbf{89.9}} & 5.8 & 10.2 & 82.0 & 9.8 & 17.7 & 78.0 & 8.3 & 14.3 & 83.3 \\
SceneTracker + ZoeDepth & - & 15.1 & - & - & 5.6 & - & - & 16.3 & - & - & 12.3 & - \\

DELTA + ZoeDepth & \cellcolor{custom_green}{\textbf{10.1}} & \cellcolor{custom_green!30}{\textbf{16.2}} & 84.7 & \cellcolor{custom_green!30}{7.8} & \cellcolor{custom_green!30}{12.8} & \cellcolor{custom_green!30}{87.2} & 10.2 & 17.8 & 74.5 & \cellcolor{custom_green}{\textbf{9.4}} & \cellcolor{custom_green!30}{15.6} & 82.1 \\
\cdashline{1-13}
Ours + ZoeDepth & \cellcolor{custom_green!30}{10.0} & 15.9 & 86.5 & \cellcolor{custom_green}{\textbf{8.0}} & \cellcolor{custom_green}{\textbf{13.5}} & \cellcolor{custom_green}{\textbf{87.3}} & 9.7 & 18.0 & 73.6 & \cellcolor{custom_green!30}{9.2} & \cellcolor{custom_green}{\textbf{15.8}} & 81.5 \\
\bottomrule
\end{tabular}
\label{tab:3d_tracking_results_zoedepth}
\end{table}

\begin{table}
\centering
\caption{\textbf{2D tracking results} on the TAP-Vid2D Benchmark. }
\setlength{\tabcolsep}{3pt}
\begin{tabular}{lccc|ccc|ccc}
\toprule
\multirow{2}{*}{\textbf{Methods}} & \multicolumn{3}{c|}{\textbf{Kinetics}} & \multicolumn{3}{c|}{\textbf{DAVIS}} & \multicolumn{3}{c}{\textbf{RGB-Stacking} }  \\

 & AJ$\uparrow$ & APD$_{2D}\uparrow$ & OA$\uparrow$ & AJ$\uparrow$ & APD$_{2D}\uparrow$ & OA$\uparrow$ & AJ$\uparrow$ & APD$_{2D}\uparrow$ & OA$\uparrow$ \\
\cdashline{1-10}
TAP-Net \citep{tapvid} & 38.5 & 54.4 & 80.6 & 33.0 & 48.6 & 78.8 & 54.6 & 68.3 & 87.7 \\
MFT \citep{mft}  &  39.6 & 60.4 & 72.7 & 47.3 & 66.8 & 77.8 & - & - & - \\
PIPs \citep{pips} & 31.7 & 53.7 & 72.9 & 42.2 & 64.8 & 77.7 & 15.7 & 28.4 & 77.1 \\
OmniMotion \citep{omnimotion} & - & - & - & 46.4 & 62.7 & 85.3 & 69.5 & 82.5 & 90.3 \\
TAPIR \citep{tapir} & 49.6 & 64.2 & 85.0 & 56.2 & 70.0 & 86.5 & 54.2 & 69.8 & 84.4 \\ 
CoTracker2 \citep{cotracker} & 48.7 & 64.3 & 86.5 & 60.6 & 75.4 & 89.3 & 63.1 & 77.0 & 87.8 \\
DOT \citep{dot} &  48.4 & 63.8 & 85.2 & 60.1 & 74.5 & 89.0 & \textbf{77.1} & \textbf{87.7} & \textbf{93.3} \\
BootsTAPIR \citep{bootstap} & \textbf{54.6} & \textbf{68.4} & 86.5 & 61.4 & 73.6 & 88.7 & 70.8 & 83.0 & 89.9\\
TAPTR \citep{taptr} & 49.0 & 64.4 & 85.2 & 63.0 & 76.1 & 91.1 & - & - & - \\
TAPTRv2 \citep{taptrv2} & 49.7 & 64.2 & 85.7 & 63.5 & 75.9 & \textbf{91.4} & - & - & -\\
LocoTrack \citep{cho2024local} & 52.9 & 66.8 & 85.3 & 63.0 & 75.3 & 87.2 & 69.0 & 83.2 & 89.5 \\
\cdashline{1-10}
SpatialTracker \citep{spatialtracker} & 50.1 & 65.9 & \textbf{86.9} & 61.1 & 76.3 & 89.5 & 63.5 & 77.6 & 88.2 \\
SceneTracker \citep{scenetracker}  & - & 66.5 & - & - & 71.8 & - & - & 73.3 & -  \\
DOT-3D & 48.1 & 63.7 & 85.9 & 61.2 & 75.3 & 88.1 & 76.3 & 86.6 & 92.1 \\
DELTA & 49.5 & 63.3 & 82.2 & 62.7 & 76.7 & 88.2 & 74.2 & 83.5 & 90.0  \\
\cdashline{1-10}
Ours  & 47.8 & 61.3 & 82.0 & 62.5 & 76.2 & 87.9 & 74.0 & 82.9 & 89.3 \\
\bottomrule
\end{tabular}
\label{tab:2d_tracking_results}
\end{table}

\begin{figure}[t]
\begin{center}
\includegraphics[width=0.96\linewidth]{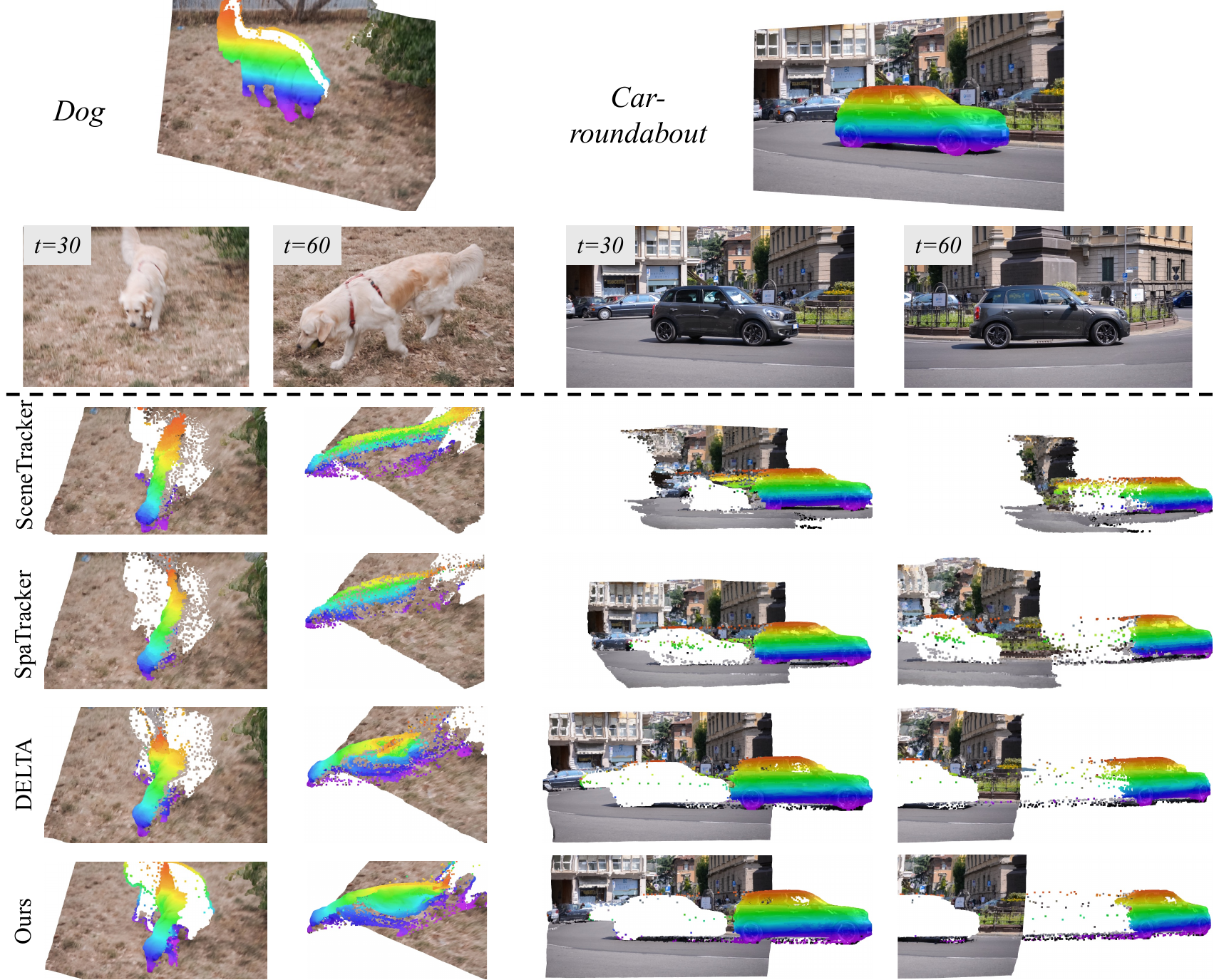}
\captionof{figure}{
   \textbf{Qualitative comparison of dense 3D tracking} between SceneTracker \cite{scenetracker}, SpaTracker \cite{spatialtracker}, DELTA \cite{ngo2024delta}, and our method. We track every pixel from the first frame through 3D space, with moving objects highlighted in \textcolor{red}{r}\textcolor{orange}{a}\textcolor{yellow}{i}\textcolor{green}{n}\textcolor{blue}{b}\textcolor{violet}{o}\textcolor{red}{w} colors. Our approach accurately captures foreground motion, preserves stable backgrounds, and operates significantly faster with a $5\times$ to $100\times$ speedup compared to prior methods. 
}
\label{fig:quali_results}
\end{center}
\end{figure}





\subsection{Dense 3D Tracking}
Figure~\ref{fig:quali_results} shows visual comparisons of dense 3D tracking between our method and baselines. Our model produces highly stable and accurate trajectories, clearly outperforming SceneTracker \cite{scenetracker} and SpaTracker \cite{spatialtracker}, and achieving results on par with the strong DELTA~\cite{ngo2024delta} baseline.


\section{Analysis of Coarse-to-fine Dense Tracking}
\subsection{Further Analysis of Strategies for Accelerating Tracking}

\begin{figure}[t]
\begin{center}
\includegraphics[width=0.96\linewidth]{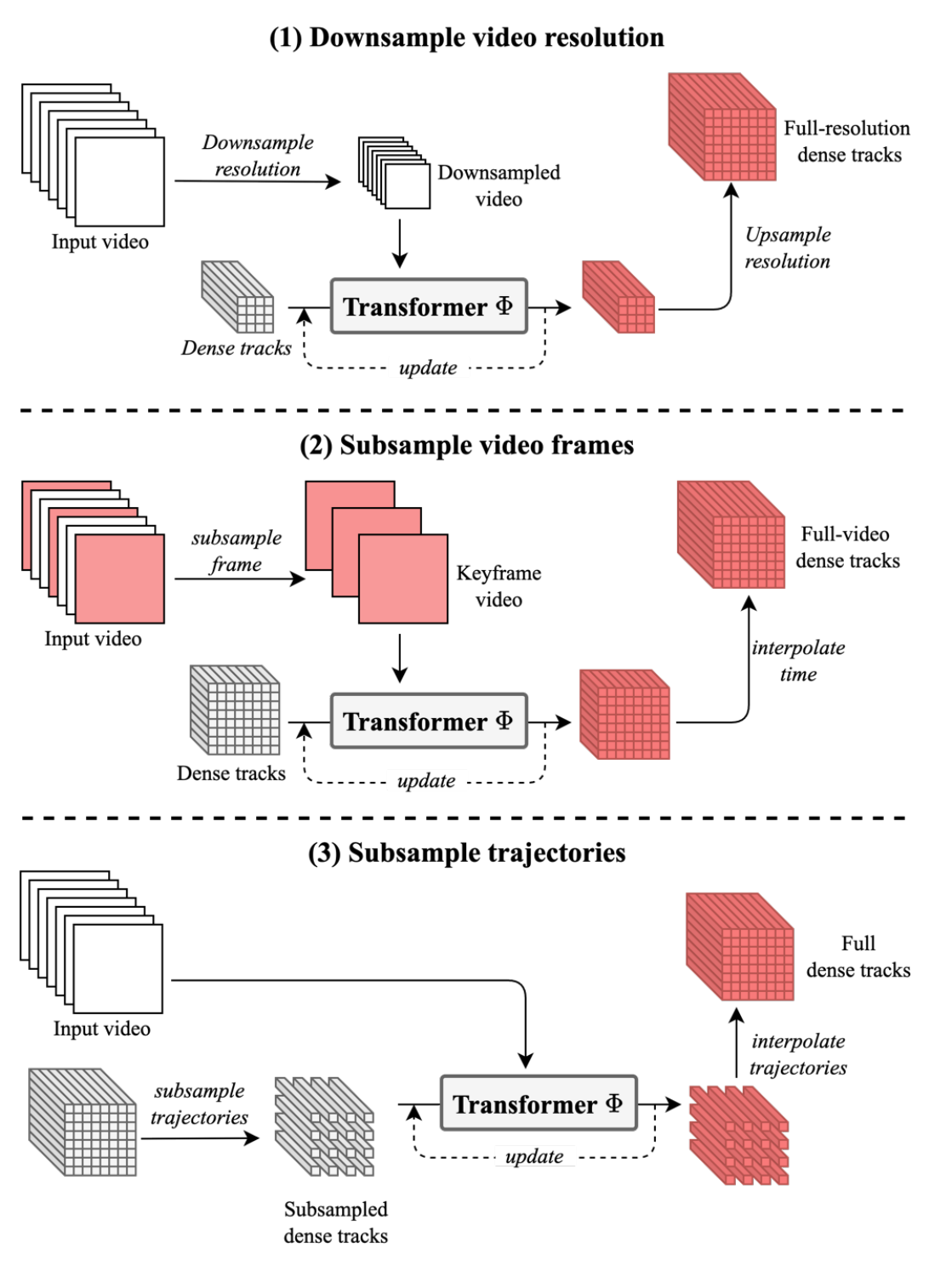}
\captionof{figure}{
   Illustration of three acceleration strategies. (1): Spatial downsampling of RGB input; (2) Temporal subsampling of input ; (3) Trajectory point subsampling;
}
\label{fig:strategies}
\end{center}
\end{figure}

Fig.~\ref{fig:strategies} illustrates the conceptual overview of each acceleration strategy described in Sec.~3.2 of the main paper. To better understand their practical impact, we further evaluate these strategies across a broader range of hyperparameters:

\noindent\textbf{1) Spatial downsampling of RGB input:} applied with spatial reduction factors of 2$\times$, 4$\times$, and 8$\times$.

\noindent\textbf{2) Temporal subsampling of input frames:} applied with temporal reduction factors of 2$\times$, 4$\times$, and 8$\times$.

\noindent\textbf{3) Trajectory point subsampling:} applied with trajectory subsampling ratios of 4$\times$, 16$\times$, and 64$\times$.

Each variant is followed by upsampling to the original spatial and temporal resolution using either (1) bilinear interpolation or (2) nearest-neighbor interpolation. The corresponding quantitative results are summarized in Tab.~\ref{tab:runtime_strategy_combined_supp}.

\subsection{Spectral Analysis of the optical flow predictions }

To analyze the spatial frequency characteristics of optical flow, we follow prior work \citep{skorokhodov2025improving} on spectral analysis using the two-dimensional discrete cosine transform (2D DCT) \citep{ahmed2006discrete}. Given a flow field $\mathcal{F}_{flow} \in \mathbb{R}^{H \times W \times 2}$, we process the horizontal and vertical components separately by dividing each into non-overlapping blocks of size $B \times B$. We then apply the type-II 2D DCT to each block, which projects the flow values onto a set of cosine basis functions at different spatial frequencies. Within each block, we arrange the coefficients into a one-dimensional sequence using the standard JPEG zigzag ordering \citep{wallace1991jpeg}, which traverses from low to high spatial frequencies. To summarize the overall spectral content of the flow, we average these zigzag-ordered coefficients across all blocks and across both flow components (horizontal and vertical). This produces a compact frequency profile that captures the distribution of motion energy across spatial scales. 

We apply this process independently to the predicted optical flow at each iteration of the model, as well as to the ground-truth flow. This allows us to track how the spectral properties of the predicted flow evolve during refinement and how they compare to the underlying ground-truth signal.

\begin{figure}[t]
  \centering
  \subfigure[W/o Coarse-to-fine]{\includegraphics[width=0.48\linewidth]{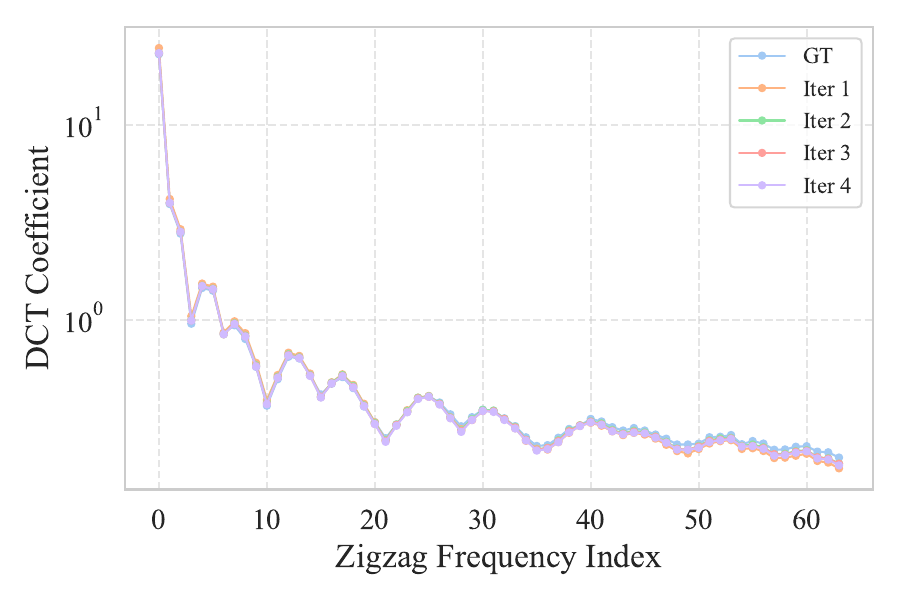} \label{fig:first}}
  \hfill
  \subfigure[With Coarse-to-fine]{\includegraphics[width=0.48\linewidth]{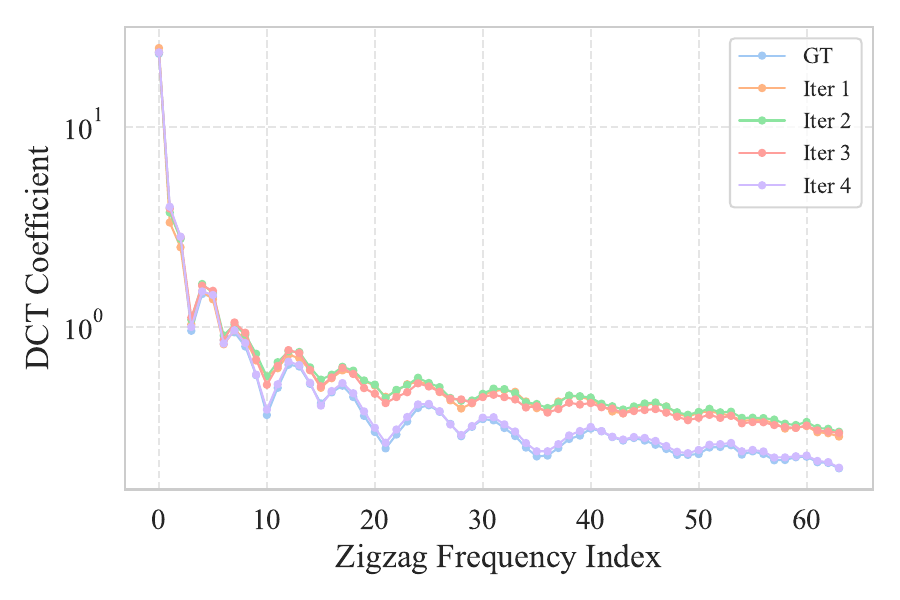} \label{fig:second}}
  \caption{DCT Spectrum of the predicted optical flows at every iteration and groundtruth between (a) w/o coarse-to-fine and (b) with coarse-to-fine.}
  \label{fig:dct_spectrum}
\end{figure}

The DCT spectrum is shown in Fig.~\ref{fig:dct_spectrum}, revealing a trade-off introduced by the coarse-to-fine strategy. While this approach helps accelerate inference and preserve low-frequency motion structures, it tends to suppress high-frequency components, which are essential for capturing fine-grained motion details and sharp flow boundaries. In contrast, the model without coarse-to-fine better preserves high-frequency content and more closely matches the spectral distribution of the ground-truth flow across all frequency bands. Notably, this limitation of the coarse-to-fine model is largely mitigated in the final iteration, where full-resolution trajectories are used. As shown in the left plot, the spectrum of the last iteration closely aligns with the ground truth, indicating that fine details can still be recovered at the final refinement stage.

\begin{figure}[t]
\begin{center}
\includegraphics[width=0.96\linewidth]{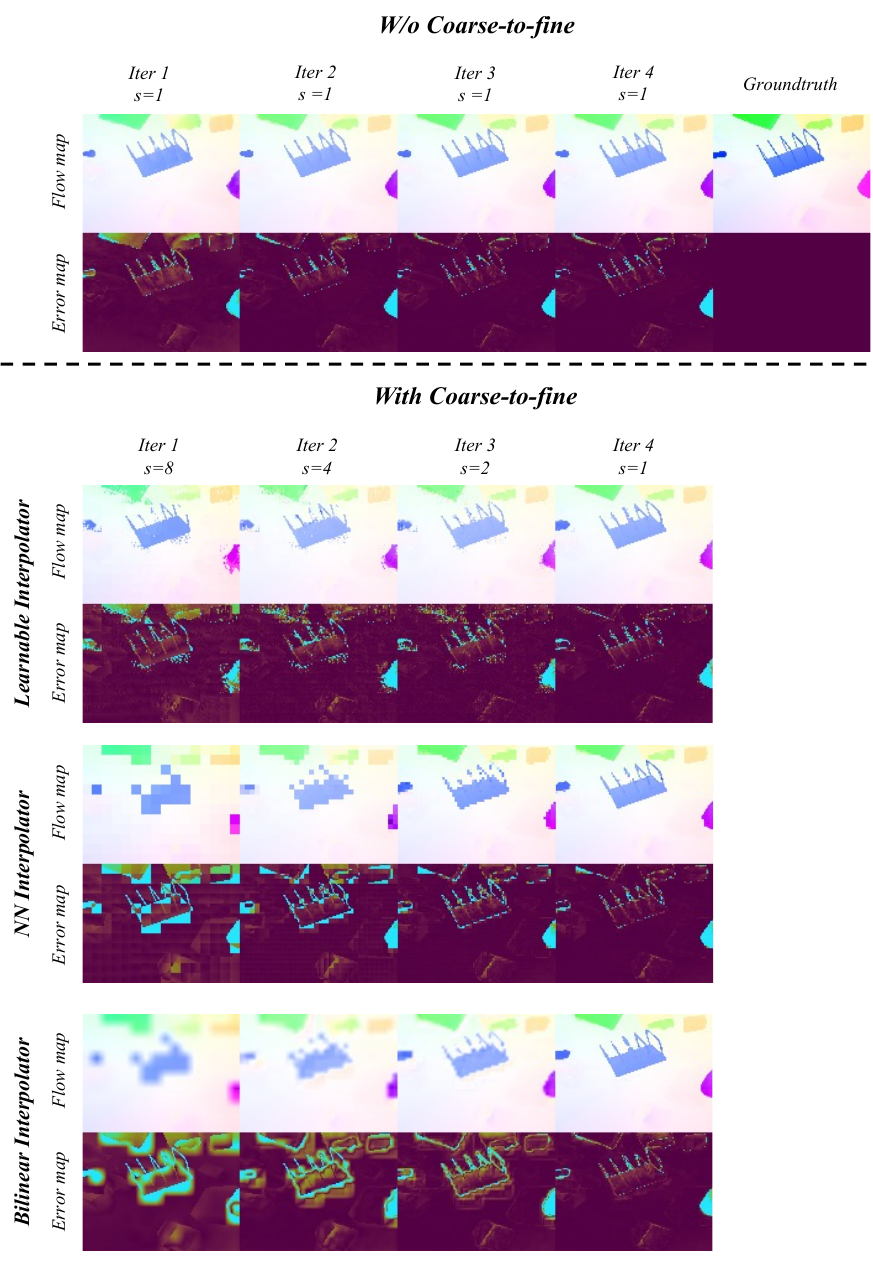}
\captionof{figure}{
   Visualization of predicted optical flow and corresponding error maps across different iterations. We compare models with and without the coarse-to-fine strategy, using various interpolation methods. 
   }
\label{fig:flow_error_iters_1}
\end{center}
\end{figure}

\begin{figure}[t]
\begin{center}
\includegraphics[width=0.96\linewidth]{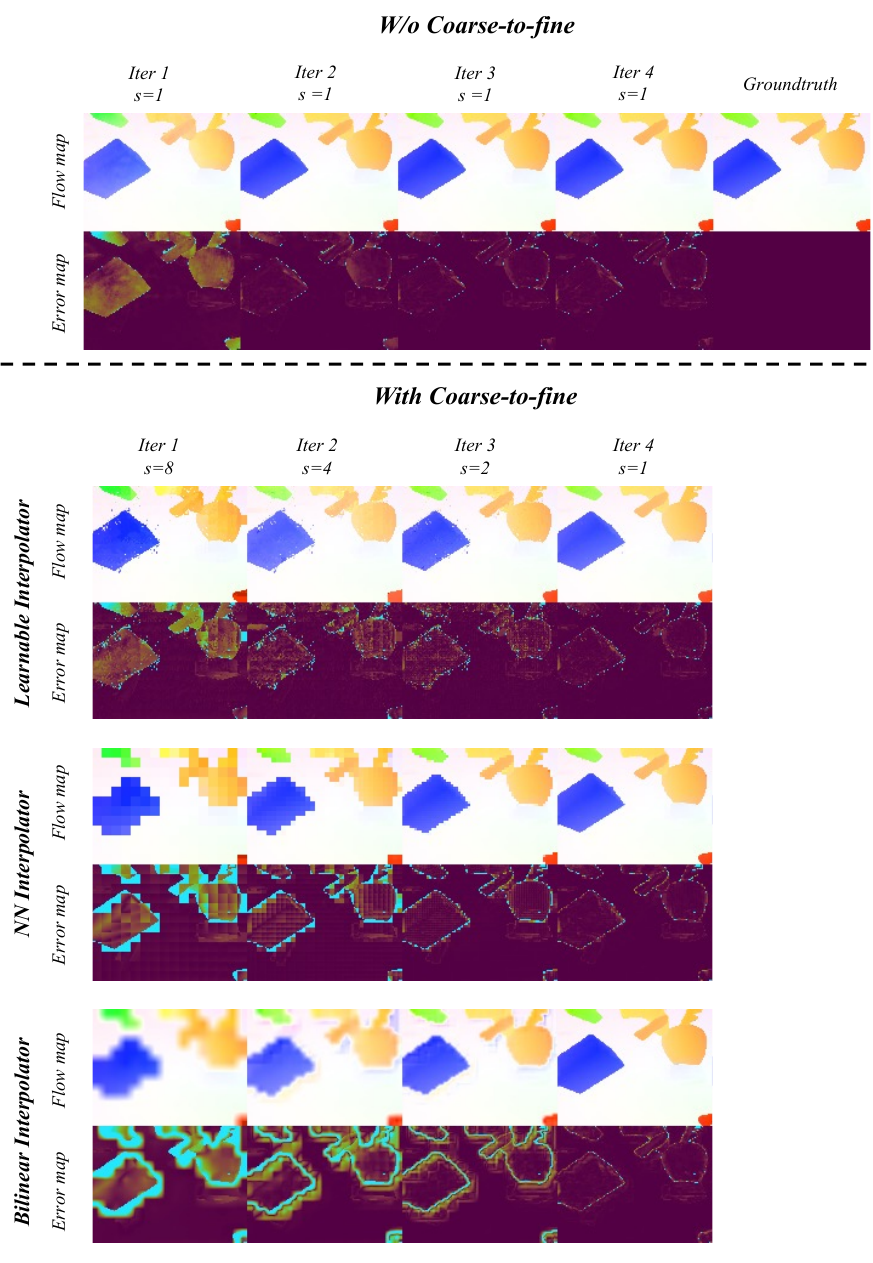}
\captionof{figure}{
   Visualization of predicted optical flow and corresponding error maps across different iterations. We compare models with and without the coarse-to-fine strategy, using various interpolation methods. 
   }
\label{fig:flow_error_iters_2}
\end{center}
\end{figure}

\begin{figure}[t]
\begin{center}
\includegraphics[width=0.96\linewidth]{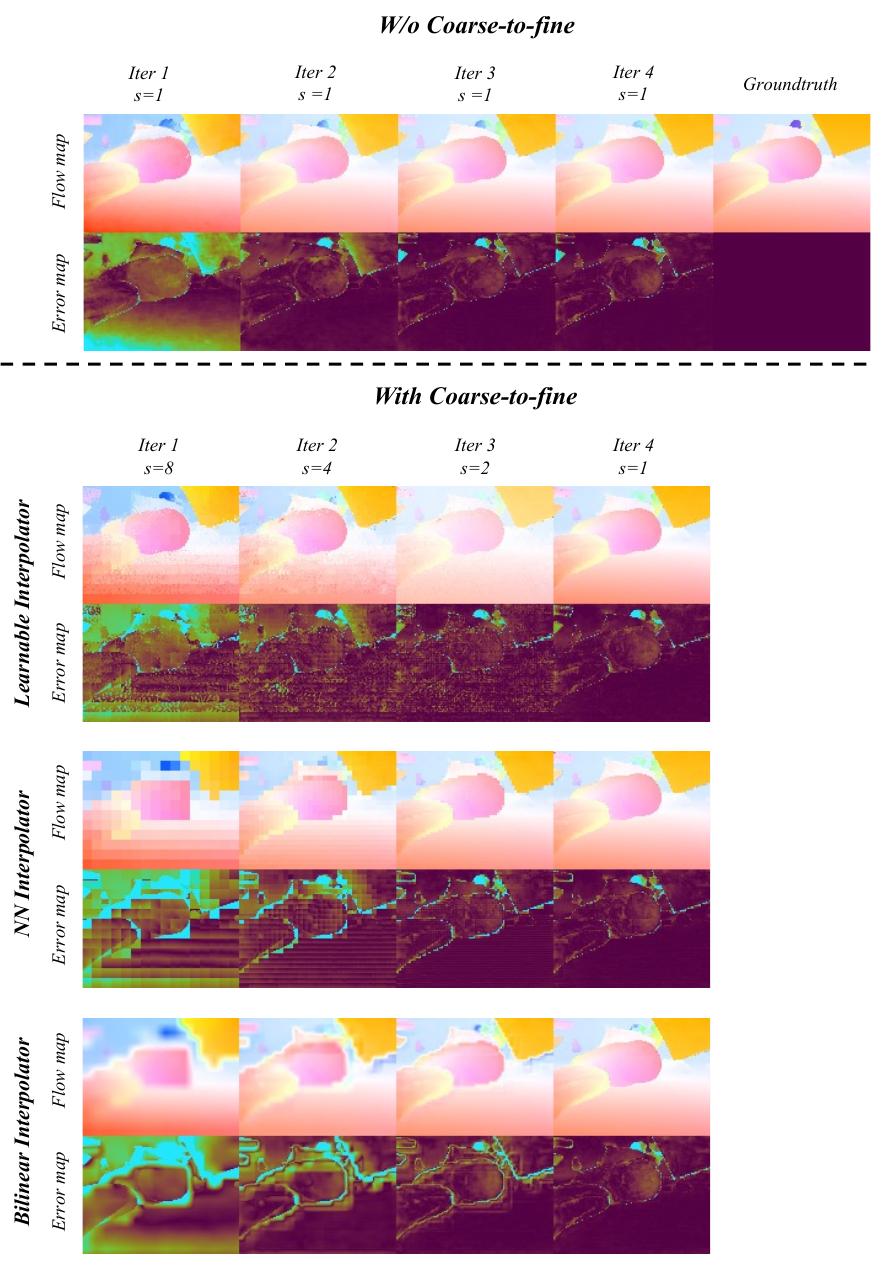}
\captionof{figure}{
   Visualization of predicted optical flow and corresponding error maps across different iterations. We compare models with and without the coarse-to-fine strategy, using various interpolation methods. 
   }
\label{fig:flow_error_iters_3}
\end{center}
\end{figure}

\begin{figure}[t]
\begin{center}
\includegraphics[width=0.96\linewidth]{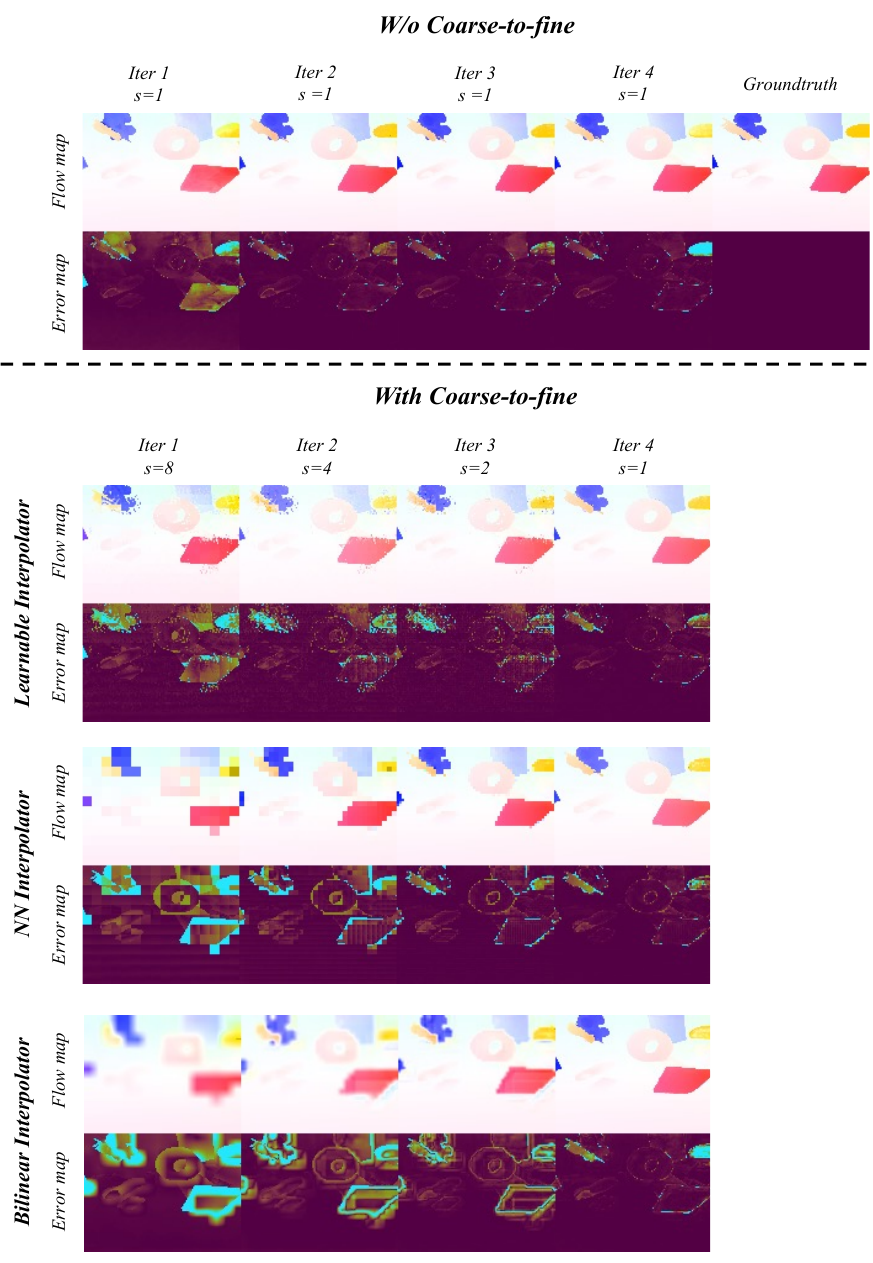}
\captionof{figure}{
   Visualization of predicted optical flow and corresponding error maps across different iterations. We compare models with and without the coarse-to-fine strategy, using various interpolation methods. 
   }
\label{fig:flow_error_iters_4}
\end{center}
\end{figure}

We visualize the predicted optical flow and corresponding error maps across different transformer iterations in Fig.~\ref{fig:flow_error_iters_1},\ref{fig:flow_error_iters_2},\ref{fig:flow_error_iters_3},\ref{fig:flow_error_iters_4}, comparing models with and without the coarse-to-fine strategy and using different interpolation methods. The results show that nearest-neighbor interpolation introduces noticeable grid artifacts, while bilinear interpolation tends to oversmooth motion boundaries. In contrast, our proposed learnable interpolator yields more accurate and spatially coherent predictions, particularly in early iterations where only a sparse grid of points is tracked and interpolation plays a critical role in producing dense motion estimates.

\begin{table}[t]
    \centering
    \small
    \setlength{\tabcolsep}{4pt}
    \caption{Comparison of cost reduction strategies on the Kubric3D \textit{val} set~\cite{ngo2024delta}. All methods use 1 iteration unless noted; full-resolution outputs are obtained using either bilinear or nearest-neighbor interpolation.}
    \begin{tabular}{lcccccc}
        \toprule
        \textbf{Baseline} & \multicolumn{3}{c}{APD$_{3D}\uparrow$} & \multicolumn{3}{c}{Runtime$\downarrow$} \\
        \midrule
        {\color{gray} DELTA (4 iterations)} & \multicolumn{3}{c}{\color{gray}{87.3}} & \multicolumn{3}{c}{\color{gray}{8404}}  \\
        \color{gray}{DELTA (1 iteration)} & \multicolumn{3}{c}{\color{gray}{74.3}} & \multicolumn{3}{c}{\color{gray}{2275}}  \\
        \midrule
        \textbf{Strategy} & \multicolumn{2}{c}{$2\times$} & \multicolumn{2}{c}{$4\times$} & \multicolumn{2}{c}{$8\times$} \\
        & APD$_{3D}\uparrow$ & Runtime$\downarrow$ & APD$_{3D}\uparrow$ & Runtime$\downarrow$ & APD$_{3D}\uparrow$ & Runtime$\downarrow$ \\
        \midrule
        \rowcolor{gray!10}
        \multicolumn{7}{l}{\textbf{(a) Bilinear interpolation}} \\

        Downsample video reso. & 54.7 & 759 & 35.8 & 383 & 17.8 & 214 \\
        Subsample video frames & 70.8 & 1119 & 65.4 & 833 & 42.3 & 491 \\
        Subsample trajectories & 72.9 & 917 & 71.2 & 585 & 68.5 & 406 \\
        \midrule
        \rowcolor{gray!10}
        \multicolumn{7}{l}{\textbf{(b) Nearest-neighbor interpolation}} \\
        Downsample video reso. & 53.9 & 748 & 34.4 & 382 & 17.6 & 212 \\
        Subsample video frames & 68.8 & 1187 & 60.0 & 821 & 49.3 & 487 \\
        Subsample trajectories & 73.7 & 902 & 72.1 & 580 & 67.6 & 405 \\
        \bottomrule
    \end{tabular}
    \label{tab:runtime_strategy_combined_supp}
\end{table}

\end{document}